%% file: main.tex
\documentclass[10pt,twocolumn,letterpaper]{article}

\usepackage[pagenumbers,final]{cvpr} 

\input{preamble}
\definecolor{cvprblue}{rgb}{0.21,0.49,0.74}
\usepackage[pagebackref,breaklinks,colorlinks,allcolors=cvprblue]{hyperref}

\usepackage{amsmath} 
\usepackage{lato} 
\usepackage{multirow}
\usepackage{tabularx}
\usepackage{setspace}
\usepackage{hhline}
\usepackage{algorithm}
\usepackage{algpseudocode}
\usepackage[accsupp]{axessibility}
\usepackage{tcolorbox}
\usepackage{listings}
\tcbuselibrary{listings, breakable}
\usepackage{ragged2e}
\usepackage[T1]{fontenc}
\usepackage{tablefootnote}   
\newcolumntype{Y}{>{\centering\arraybackslash}X}
\newcolumntype{U}{>{\raggedleft\arraybackslash}X}
\newcolumntype{L}{>{\raggedright\arraybackslash}X}
\usepackage[table]{xcolor}
\newtcblisting{promptbox}{
  breakable,
  listing only,
  listing engine=listings,
  colback=gray!5,
  colframe=gray!40,
  boxsep=0.5mm,    
  left=0.5mm,        
  right=0.5mm,       
  top=1mm,
  bottom=1mm,
  listing options={
    basicstyle=\scriptsize\ttfamily,
    breaklines=true,          
    breakatwhitespace=false,  
    columns=fullflexible,     
    xleftmargin=0pt,          
    framexleftmargin=0pt,     
    breakindent=0pt           
  },
}

\def\paperID{22566} 
\def\confName{CVPR}
\def\confYear{2026}

\usepackage{multirow}

\title{Agile Deliberation: Concept Deliberation for Subjective Visual Classification
\vspace{-1ex}
}

\author{
Leijie Wang$^{\dagger,\S}$\thanks{This work was done during an internship at Google.} \quad
Otilia Stretcu$^{\dagger}$ \quad
Wei Qiao$^{\ddagger}$ \quad
Thomas Denby$^{\ddagger}$ \quad
Krishnamurthy Viswanathan$^{\dagger}$\\
Enming Luo$^{\dagger}$ \quad
Chun-Ta Lu$^{\dagger}$ \quad
Tushar Dogra$^{\ddagger}$ \quad
Ranjay Krishna$^{\S}$ \quad
Ariel Fuxman$^{\dagger}$\\[0.5em]
$^{\dagger}$Google Research \quad
$^{\ddagger}$Google \quad
$^{\S}$University of Washington\\
{\tt\small leijiew@cs.washington.edu \quad otiliastr@google.com}
\vspace{-1ex}
}

\usepackage{minted}
\begin{document}
\maketitle
\vspace{-6mm}

\begin{figure*}[t!]
    \centering
    \vspace{-1em}
    \includegraphics[width=0.94\textwidth]{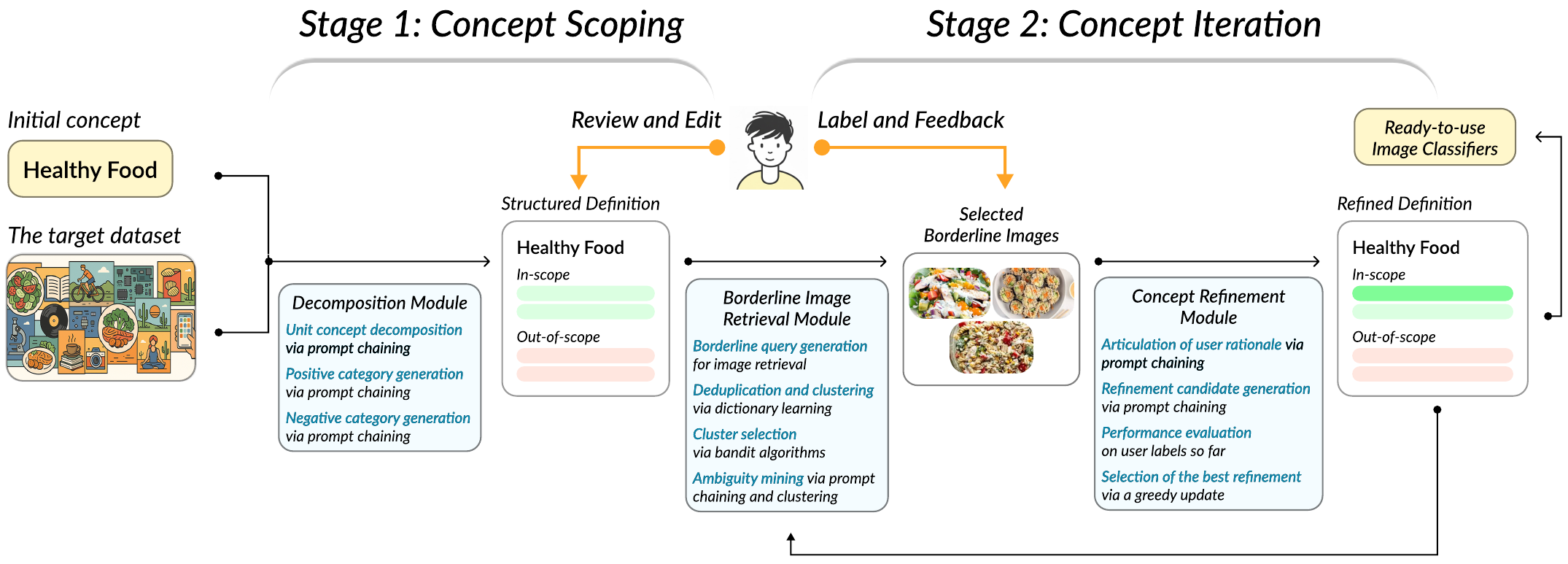}
    \caption{\textbf{Overview of the Agile Deliberation framework and architecture}. Given a subjective concept and a target dataset, \textit{Agile Deliberation} produces both a structured concept definition and an image classifier through a human-in-the-loop deliberation process.
    At the \textit{scoping stage}, the \textit{decomposition module} helps users break down their initial concept into a structured definition.
    At the \textit{iteration stage}, the \textit{borderline image retrieval module} surfaces semantically borderline images for user labeling and reflection, while the \textit{concept refinement module} updates the definition to incorporate this feedback and preserve performance on previously labeled data.}
    \label{system_architecture}
    \vspace{-1em}
\end{figure*}

\input{00_abstract}

\input{01_introduction}

\input{02_related_work}

\input{03_method}
\input{04_experiments}

\input{06_conclusion}

\section{Acknowledgements}
We would like to thank members of the Agile Modeling team at Google and of Social Futures Lab at the University of Washington for their invaluable help in this project. We also would like to thank our anonymous reviewers for their insightful feedback. Finally, we would like to express our heartfelt thanks to all the participants who dedicated their time and effort to participate in our study.

{
    \small
    \bibliographystyle{ieeenat_fullname}
    \bibliography{main}
}

\clearpage
\input{appendix}
\end{document}

%% file: 00_abstract.tex
\begin{abstract}

From content moderation to content curation, applications requiring vision classifiers for visual concepts are rapidly expanding.
Existing human-in-the-loop approaches typically assume users begin with a clear, stable concept understanding to be able to provide high-quality supervision.
In reality, users often start with a vague idea and must iteratively refine it through ``concept deliberation'', a practice we uncovered through structured interviews with content moderation experts.
We operationalize the common strategies in deliberation used by real content moderators into a human-in-the-loop framework called \textit{Agile Deliberation} that explicitly supports evolving and subjective concepts.
The system supports users in defining the concept for themselves by exposing them to borderline cases. The system does this with two deliberation stages: (1) concept scoping, which decomposes the initial concept into a structured hierarchy of sub-concepts, and (2) concept iteration, which surfaces semantically borderline examples for user reflection and feedback to iteratively align an image classifier with the user’s evolving intent. 
Since concept deliberation is inherently subjective and interactive, we painstakingly evaluate the framework through 18 user sessions, each 1.5h long, rather than standard benchmarking datasets.
We find that Agile Deliberation achieves 7.5\% higher $F_1$ scores than automated decomposition baselines and more than 3\% higher than manual deliberation, while participants reported clearer conceptual understanding and lower cognitive effort.

\end{abstract}

%% file: 01_introduction.tex
\vspace{-1ex}
\section{Introduction}
Computer vision as a field has traditionally focused on recognizing concepts that are objectively agreed upon, such as dogs, cars, or tomatoes~\cite{russakovsky2015imagenet}.
However, many real-world vision applications increasingly demand recognizing subjective concepts whose boundaries are ill-defined and often contested~\cite{kiela2020hateful, toubal2024modeling}. 
For example, content moderators may disagree on what qualifies as unsafe imagery, while food critics may hold distinct interpretations of what visually represents gourmet food.
Traditional pipelines (e.g., crowdsourced labeling~\cite{kovashka2016crowdsourcing} and fixed taxonomies~\cite{ibrahim2023explainable}) presume a single, well-defined ground truth and thus struggle to capture such conceptual variability~\cite{stretcu2023agile}.

Many human-in-the-loop methods have thus begun to address this gap.
For instance, Agile Modeling~\cite{stretcu2023agile} enables users to rapidly bootstrap vision classifiers for subjective concepts by iteratively labeling hundreds of training images.
With the surge of large language models (LLMs), another increasingly prevalent approach is for domain experts to write custom prompts for visual language models (VLMs) ~\citep[e.g.,][]{qiao2024scaling}.

However, these methods typically assume that the concept definition is \textit{static and well-understood by the user}.
In practice, people often start with a vague understanding of their subjective concepts and must spend significant time exploring borderline cases and refining their definitions accordingly~\cite{ma2025should, wang2025end}.
Such concept deliberation is crucial for producing high-quality supervision to downstream classifiers.
Without a carefully articulated definition as few-shot prompts, downstream VLM-based classifiers may resolve ambiguities arbitrarily and fail to capture intended decision boundaries~\cite{zamfirescu2023johnny, wang2025end, ma2025should}.
Moreover, even experts can produce inconsistent annotations as their concept understanding evolves with increased exposure to examples~\cite{kulesza2014structured, pandey2022modeling}.
Such inconsistencies are particularly harmful when only small datasets are available for fine-tuning~\cite{klie2023annotation, mullen2019comparing, pang2025token}, as in most real-world applications like content moderation~\cite{wang2025end}.

In this paper, we propose \textit{Agile Deliberation}, an interactive human-in-the-loop framework that explicitly supports evolving, subjective concepts\footnote{Agile Deliberation is available at \url{https://github.com/google-research/google-research/tree/master/agile_deliberation}}.
It helps users articulate a \textit{concept definition} that serves as both a human-readable description and a textual prompt for VLMs to derive a performant image classifier.
Our framework was inspired by structured interviews with real content moderation experts. We also analyze expert-authored concept definitions, from which we reveal common strategies that real users employ to tackle the challenges in deliberating over subjective concepts.
Based on these insights, \textit{Agile Deliberation} structures concept deliberation into two main stages: \textbf{concept scoping} and \textbf{concept iteration}.
At the scoping stage, a \textit{decomposition module} uses prompt-chained reasoning~\cite{zhou2022least} to expand the user’s initial concept into a hierarchy of subconcepts, mirroring how humans apply compositional logic to clarify meaning~\cite{miller1956magical, frege1879begriffsschrift}.
At the iteration stage, a \textit{borderline image retrieval module} surfaces \textit{semantically borderline examples} near the decision boundary of current definition for user labeling and reflection.
We specifically target \textit{semantically} borderline examples rather than rely on standard active learning methods~\cite{stretcu2023agile} because VLM predictions are often uncalibrated with human understanding~\cite{mendes2025uncertainty, tomov2025illusion}.
A \textit{concept refinement module} then integrates this feedback by refining the definition automatically~\cite{Pryzant2023AutomaticPO}.
By iterating between concept decomposition, example-based deliberation, and prompt optimization, our method steadily aligns the image classifier with user intent.

Our second contribution lies in evaluating this framework through $18$ sessions with non-expert users, each $1.5h$ long.
Because subjective concepts lack a static ground truth, we depart from standard offline benchmarks which assumes fixed definitions. Instead, we have to validate our framework through live user sessions, directly measuring the system’s ability to support the dynamic evolution of user intent—a process static datasets cannot capture.
Compared with automated baselines without concept deliberation, our method achieved an average improvement of 10.5\% in $F_1$ scores over zero-shot classifiers~\cite{chen2022pali} and 7.5\% over methods that automatically decompose subjective concepts using LLMs~\cite{toubal2024modeling}.
It also outperformed settings where participants conducted manual deliberation with expected baseline supports.
Our analysis of participant feedback suggests that these performance gains resulted from clearer concept understanding and better human-VLM alignment.
We also found that \textit{Agile Deliberation} makes defining and training subjective classifiers accessible beyond experts, allowing more people to easily encode their values and perspectives into vision systems.
By foregrounding concept deliberation, it also empowers practitioners in high-stakes domains like content moderation to proactively design classifiers that anticipate nuanced risks and harms, rather than merely responding to them.

%% file: 02_related_work.tex
\section{Related Work}

\textbf{Human-in-the-Loop Visual Classification}.
Traditional visual recognition pipelines typically assume a well-defined ground truth via crowdsourced labeling~\citep{kovashka2016crowdsourcing} or fixed taxonomies~\citep{ibrahim2023explainable}, struggling to capture subjective concepts.
To address this, various human-in-the-loop paradigms have been developed~\cite{Branson2010VisualRW, Ratner2017SnorkelRT, koh2020concept}. 
Closer to our work, Agile Modeling~\citep{stretcu2023agile} and Modeling Collaborator~\citep{toubal2024modeling} let users rapidly bootstrap classifiers for subjective concepts through iterative manual labeling or automated LLM-based decomposition.
However, they generally assume that concept definitions are static and well-understood from the start.
Our work instead models the evolving nature of subjective concepts and integrating deliberative feedback directly into the classifier training loop.

\vspace{1ex}
\textbf{Zero and Few-shot Learning of Vision-language Models (VLMs).}
Recent advances in vision–language models~\citep[e.g.,][]{radford2021learning,jia2021scaling,Li2022BLIPBL} demonstrate strong generalization from minimal supervision, yet they rely heavily on well-specified, static prompts.
We build on these foundations by introducing a dynamic process that iteratively updating the prompts to align with evolving user intent.

\vspace{1ex}
\textbf{Borderline Example Discovery and Active Learning}.
Active learning and uncertainty sampling methods select informative samples, typically near the decision boundary, to maximize performance improvement with minimal labeling~\citep{Settles2009ActiveLL, lewis1995sequential}.
Our method extends this idea by surfacing \textit{semantically} borderline examples that align with users’ conceptual ambiguities.
Unlike standard uncertainty sampling, it organizes borderline samples along interpretable dimensions to provoke human reflection and boundary refinement. 

\vspace{1ex}
\textbf{Prompt Optimization and Tuning.}
Research on improving VLM performance generally falls into two categories: prompt tuning, which learns continuous soft prompts via backpropagation~\citep{Lester2021ThePO,zhou2022learning}, and automatic prompt optimization (APO), which searches for improved discrete textual prompts~\citep{agrawal2025gepareflectivepromptevolution, prasad2023grips, Pryzant2023AutomaticPO, Zhou2022LargeLM}. While effective, these methods typically optimize a fixed objective on a static validation set. We adapt APO paradigms to a human-in-the-loop setting, using rich, deliberative feedback—rather than scalar metrics—to progressively align the prompt with human intent.

%% file: 03_method.tex
\section{Agile Deliberation}
\label{sec:method}

We propose {\em Agile Deliberation}, a human-in-the-loop framework for visual classification that integrates concept deliberation into the classifier training pipeline.
Formally, let $\mathcal{X}$ denote the space of images and $s$ a user-provided \emph{subjective concept name} (e.g., $s = \texttt{healthy food}$).
Optionally, the user may provide a \emph{target dataset} of unlabeled images $\mathcal{D} = \{x_i\}_{i=1}^N \subset \mathcal{X}$, or we use a large-scale web dataset (e.g., WebLI~\cite{chen2022pali}) as the default domain.

Through iterative deliberation, the system constructs a \textit{structured concept definition} $d \in \mathcal{D}_{\text{def}}$, a textual specification with explicit positive/negative subconcepts and edge cases capturing the user’s interpretation of the concept.
This definition serves as both:
(1) A human-readable articulation of the concept to support consistent labeling, and
(2) A \textit{textual prompt} for a vision–language model (VLM), which induces a classifier
\begin{equation}
f_d : \mathcal{X} \to [0,1], \qquad f_d(x) = P(y = 1 \mid x; d)
\end{equation}
where $y \in \{0,1\}$ indicates whether an image is \textit{in-scope}, or a positive instance of the concept.

Over rounds $t=1,2,\dots,T$, Agile Deliberation incrementally expands a labeled set
\begin{equation}
    \mathcal{L}_t = \{(x_i, y_i)\}_{i=1}^{n_t} \subset \mathcal{D} \times \{0,1\}
\end{equation}
and updates the definition $d_t$ by choosing
\begin{equation}
    d_{t+1} = \arg\max_{d' \in \mathcal{C}_t} F_1\bigl(f_{d'},\, \mathcal{L}_t\bigr)
\end{equation}
where $\mathcal{C}_t \subseteq D_{def}$ is a set of candidate refined definitions generated from user feedback, and $F_1\bigl(f_{d'},\, \mathcal{L}_t\bigr)$ is the F1-score of the predictions of classifier $f_{d'}$ on $\mathcal{L}_t$.
In other words, our deliberation process \textit{optimizes definitions that serve as effective VLM prompts} under the user’s labels.
While this optimization formally targets classifier performance, the structured format of $d_t$ also encourages consistency in how users apply their definitions across rounds.

In the following sections, we first describe the overall Agile Deliberation framework and then detail our prototype implementation, specifying how each component realizes these functionalities.


\subsection{The Framework}

\begin{figure*}
    \centering
    \includegraphics[width=0.85\textwidth]{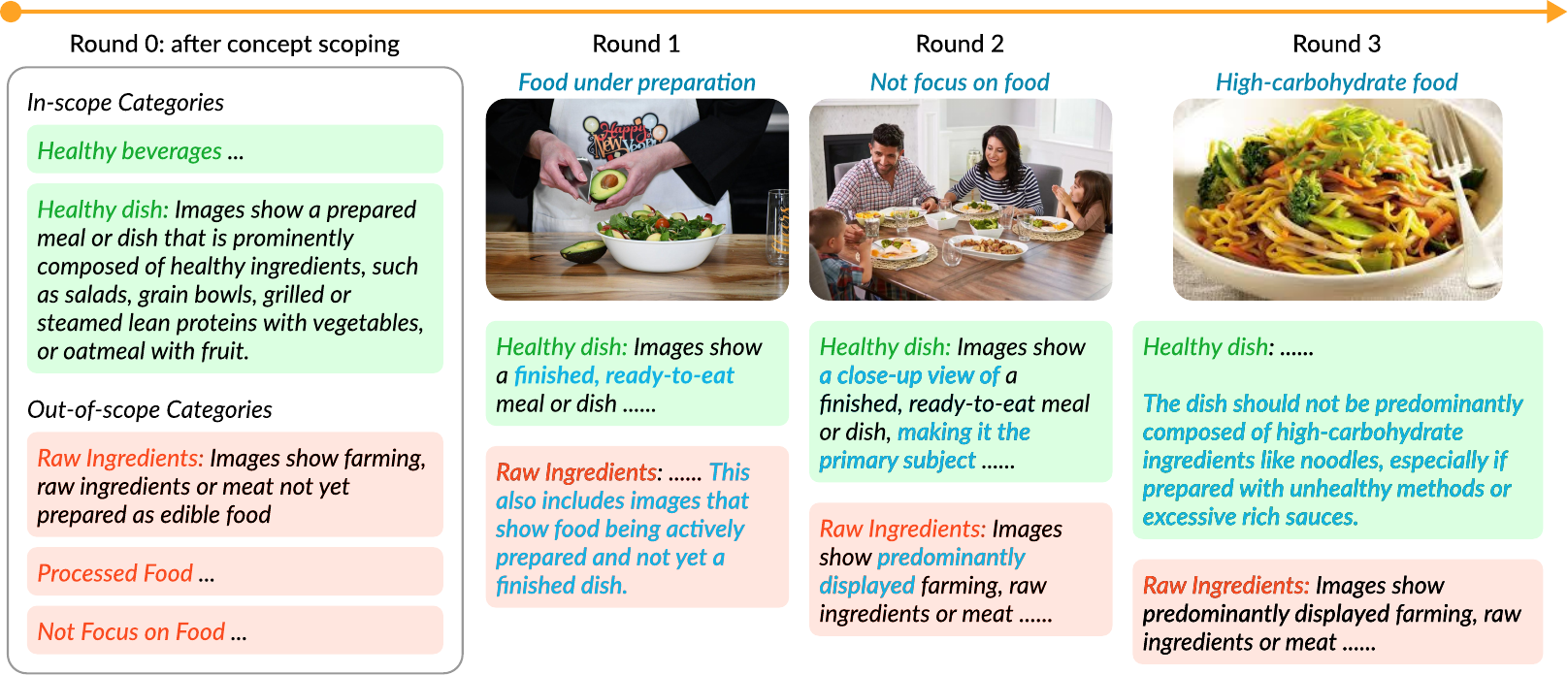}
    \caption{\textbf{Example of iterative concept refinement in \textit{Agile Deliberation}} (from an actual study participant).
    We show the first three iteration rounds, highlighting updates within two subconcepts for brevity.
    Only one representative image from each batch of borderline images is displayed for illustration.
    In the \textit{concept scoping} stage, the participant first decomposed their initial concept into in-scope and out-of-scope subconcepts.
    In the subsequent \textit{concept iteration} stage, they reviewed surfaced borderline images, provided labels and optional feedback, and the system incorporated this feedback into progressively refined definitions.}
    \hfill
    \label{concept_evolution}
    \vspace{-1em}
\end{figure*}

As illustrated in Figure~\ref{system_architecture}, Agile Deliberation consists of two stages: \textbf{(1) concept scoping} and \textbf{(2) concept iteration}.
Our design is informed by interviews with 5 content moderation experts in who routinely reason about subjective concepts, along with qualitative coding of 20 high-quality concept definitions sampled from their professional workflows.
We learned that practitioners typically begin by \textit{scoping} their concept definitions (reviewing representative images to identify key visual signals) and \textit{refine} them by searching for and reflecting on borderline images.
However, they struggle to efficiently surface borderline images and align downstream classifiers with their nuanced understanding (See Appendix \ref{interview_appendix} for full details).

Agile Deliberation operationalizes this workflow by structuring both stages within a human-in-the-loop training pipeline.
In concept scoping, the system decomposes the user’s initial concept description into a hierarchy of subconcepts.
The second stage, content iteration, involves multiple rounds where the system retrieves and selects a batch of borderline images through a structured workflow for users to reflect on and label.
Their feedback is then used to update the concept definition, progressively improving an image classifiers with the user's evolving interpretation.

For illustration, we use \texttt{healthy food} as a running example.
It captures ambiguities common in real-world content moderation though the same process applies equally to other subjective or more sensitive concepts.


\vspace{1ex}
\noindent\textbf{Stage 1:~Concept scoping.}
The process begins with a user-provided subjective concept, such as
$s = $\texttt{healthy food}.
If $s$ is a composite concept (e.g., \texttt{healthy food}), the system first decomposes it into simpler unit concepts $u_1, .., u_M$ that capture its core visual dimensions.
Each unit concept $u_m$ is then expanded through prompt-chained reasoning ~\cite{zhou2022least} into subconcepts:
\begin{equation}
    \mathcal{S}_m^+ = \{s^+_{m,1}, \dots, s^+_{m, K_+}\},
\mathcal{S}_m^- = \{s^-_{m,1}, \dots, s^-_{m,K_-}\}
\end{equation}
where $\mathcal{S}_m^+$ denotes candidate \emph{positive} subconcepts (e.g., \texttt{healthy dish}, \texttt{fresh fruit}) and $\mathcal{S}_m^-$ denotes candidate \emph{negative} subconcepts (e.g., \texttt{fried fast food}, \texttt{processed snacks}).
For each subconcept $s_{m,j} \in \mathcal{S}_m^+ \cup \mathcal{S}_m^-$, the system generates text queries and retrieves related images from the target dataset $\mathcal{D}$ for user inspection.
Users then decide whether each subconcept should be included as a positive or negative subconcept (or discarded if it does not represent meaningful visual patterns).

The result of Stage~1 is an initial structured definition $d_0 = \bigl\{\,(\mathcal{S}^+_{m},\, \mathcal{S}^-_{m})\,\bigr\}_{m=1}^{M}$.
This definition then instantiates the first VLM-based image classifier $f_{d_0}$ and seeds subsequent iterations (see Appendix \ref{definitions_appendix} for examples).



\vspace{1ex}
\noindent\textbf{Stage 2.~Concept iteration.}
After establishing the concept boundary, the system enters an iterative refinement stage indexed by $t = 1,2,\dots,T$.
At each round $t$, the framework:
(1) \textit{Retrieves borderline images} that are semantically ambiguous under $d_t$, and
(2) \textit{Updates the concept definition} $d_t$ based on user labels and feedback.
A key design choice is to target \textit{semantic} borderline examples, rather than performing classical model-based active learning in a classifier’s probability space, as in Agile Modeling~\citep{stretcu2023agile}.
Standard active learning methods such as uncertainty sampling~\cite{lewis1995sequential} select images at the classifier's decision boundary, which for binary classification is equivalent to selecting samples with prediction probabilities near $0.5$, or $\lvert f_{d_t}(x) - 0.5 \rvert \approx 0$. This assumes $f_{d_t}(x)$ is a calibrated probability and works well for improving a specific \textit{classifier} of interest, but may not be aligned with human understanding. 
In contrast, VLMs are generative models whose outputs are often uncalibrated and may assign high confidence to cases that humans find ambiguous and vice versa~\cite{mendes2025uncertainty, tomov2025illusion}.
Moreover, our ``classifier” $f_{d_t}$ is instantiated by prompting a generative VLM rather than a dedicated discriminative model, making classical margin-based sampling ill-defined.
Instead, we identify borderline regions in \textit{semantic space}.
Intuitively, we seek to discover a set of borderline images $B_t \subset D$ such that all images $x \in B_t$ are near the natural-language decision boundary implied by $d_t$, as judged by LLM/VLM reasoning rather than by model confidence.

These borderline examples stress-test the current definition along one interpretable dimension (e.g., \textit{amount of creamy sauce,} \textit{sugar content,} \textit{portion size}), prompting users to refine their criteria.
Example borderline sets may consist of images of healthy dishes combined with creamy sauces, pushing users to clarify whether such dishes remain in-scope for \texttt{healthy food}.
For each borderline image, the system also generates a short textual summary explaining the ambiguity to aid user reflection.

Users label each image $x \in \mathcal{B}_t$ as in-scope ($y=1$) or out-of-scope ($y=0$), updating the labeled set
\begin{equation}
    \mathcal{L}_{t} = \mathcal{L}_{t-1} \cup \{(x, y) : (x, y) \in \mathcal{B}_t \times \{0,1\}\}
\end{equation}
The interface juxtaposes user labels with $f_{d_t}(x)$ and its rationale.
When labels and predictions disagree, users provide brief justifications (e.g., ``\textit{too much cream in this salad; a light drizzle can be in-scope}'').
The prompt optimization module then integrates this feedback to propose new definition candidates $\mathcal{C}_t$; as described later, we choose
\begin{equation}
    d_{t+1} = \arg\max_{d' \in \mathcal{C}_t} F_1\bigl(f_{d'},\, \mathcal{L}_t\bigr)
\end{equation}
and present $d_{t+1}$ for users' optional manual edits.
Through repeated rounds, the textual definition and its induced classifier $f_{d_t}$ converge toward users' evolving understanding.


\subsection{The Prototype}
In this section, we detail the modeling choices behind each component in the Agile Deliberation framework (see the blue boxes in Figure~\ref{system_architecture}). 
We instantiate Agile Deliberation with two widely available tool classes: 1)
\textbf{Image retrieval engines} that map a text query $q$ to a set of visually similar images from the target dataset, $R(q, D) \subset D$,
implemented via web image search (e.g., Google Images) or via similarity search using image–text embedding models such as CLIP~\cite{ramesh2022hierarchical} or ALIGN~\cite{jia2021scaling}.
2) \textbf{VLM-based image classifiers} that combine chain-of-thought reasoning with textual prompts to decide whether $x$ is in-scope.
Concretely, given a definition $d$ and an image $x$, the VLM generates a rationale $r(x,d)$ and a binary decision
$f_d(x) \in \{0, 1\}$, following Modeling Collaborator~\cite{toubal2024modeling}.
Built on top of these tools, the system comprises three main components:
\begin{enumerate}
\item A \textit{decomposition module} that decomposes an initial concept into a hierarchy of subconcepts and visual modes;
\item A \textit{borderline image retrieval module} that surfaces semantically ambiguous examples for user review; and
\item A \textit{concept refinement module} that refines the concept definition based on user feedback and labeled data.
\end{enumerate}
Since effective human-in-the-loop deliberation demands real-time interactivity, we favor architectures with low latency over more computationally demanding alternatives.
See implemented prompts in Appendix \ref{prompts_appendix}.

\vspace{1ex}
\noindent\textbf{Decomposition module.}

By collaborating with domain experts, we identified a definition pattern that supports effective concept deliberation.
Building on this analysis, our system decomposes a subjective concept into a hierarchy of subconcepts using prompt-chained reasoning~\cite{zhou2022least}, echoing how users naturally combine visual ideas through first-order logic~\cite{miller1956magical, frege1879begriffsschrift}.
A user-provided concept $s$ is first decomposed into one or more \textit{unit concepts}:
\begin{equation}
    s \equiv \phi(u_1,\dots,u_M), \qquad u_m \in \mathcal{U}, M\leq3
\end{equation}
Here $\phi$ is a formula over conjunctions and disjunctions.
We keep $M$ small because too many unit concepts are harder to reason about.
Examples include \texttt{\small people exercising} ($u_1=$\texttt{\small people}, $u_2=$\texttt{\small exercises}) or \texttt{\small before and after achievements} ($u_1=$\texttt{\small before and after layout}, $u_2=$\texttt{\small achievements}).
For each unit concept $u_m$, $S^+_m$/$S^-_m$ candidate positive/negative subconcepts are generated with retrieved representative images.
Users then decide which subconcept should be included, which leads to a structured definition $d_{0}$ and an initial classifier $f_{d_0}$.

\vspace{1ex}
\noindent\textbf{Borderline image retrieval module.}
A key challenge in agile deliberation is to identify borderline images $\mathcal{B}_t$ that expose subtle decision boundaries,
are grounded in the target dataset, and
reflect user-relevant ambiguities rather than generic model uncertainties.
Directly prompting an LLM to propose borderline queries is attractive but problematic: it may overlook idiosyncratic user preferences, hallucinate non-existent visual descriptions, or produce descriptions difficult to retrieve accurately.
To address this, we adopt a structured retrieval-and-selection workflow:

\begin{enumerate}
    \item \textit{Borderline query generation}: Given the expanded definition $d_0$, we prompt an LLM to generate a diverse set of borderline queries ${q_j}_{j=1}^J$ (e.g., \texttt{\small salads with heavy mayo dressings}, \texttt{\small sweet fruit juice}).
    For each $q_j$, we retrieve
    $\mathcal{C}_j = R(q_j, \mathcal{D})$ with
    $\lvert \mathcal{C}_j \rvert \approx 50\text{--}100$, prioritizing surfacing a broader pool of borderline cases.

    \item \textit{De-duplication and clustering}: 
    We first remove near-duplicate images across $\bigcup_j \mathcal{C}_j$.  
    Each image $x$ is then represented by a feature vector $\mathbf{z}(x) \in \mathbb{R}^p$ (using a vision encoder) and apply dictionary learning~\cite{tovsic2011dictionary} to learn a basis $W \in \mathbb{R}^{p \times K}$ and sparse codes $\alpha(x) \in \mathbb{R}^K$ such that
    $\mathbf{z}(x) \approx W \alpha(x)$.
    Images with similar sparse codes are grouped into (possibly overlapping) clusters
    $\mathcal{G} = \{G_1,\dots,G_M\}, \quad G_m \subset \mathcal{D}$, each corresponding to a shared visual characteristic.

    \item \textit{Cluster selection}:
    At iteration $t$, we treat each cluster index $m \in \{1,\dots,M\}$ as an arm in a multi-armed bandit~\cite{slivkins2019introduction}.  
    When a cluster $G_m$ is shown to the user, we record a reward $r_t(m)$ summarizing its usefulness (e.g., fraction of classifier errors corrected, or amount of rich feedback).  
    We maintain an estimate $\widehat{\mu}_t(m)$ of the expected reward and select the next cluster using an upper-confidence bound (UCB) rule:
    \begin{equation}
        m_t = \arg\max_m \Big[\, \widehat{\mu}_t(m) + \beta_t \sqrt{\frac{\log t}{n_t(m)}} \,\Big]
    \end{equation}
    where $n_t(m)$ is the number of times cluster $m$ has been selected and $\beta_t$ controls the exploration–exploitation trade-off.  
    This directs user attention to clusters where the model disagrees with users or feedback has historically been most informative, but downweights already ``solved'' clusters (high agreement, low ambiguity).

    \item \textit{Ambiguity mining}:
    Even within a chosen cluster $G_{m_t}$, the cardinality $|G_{m_t}|$ can be large.  
    We therefore sample a manageable subset $g \subset G_{m_t}$ and ask a VLM to generate one-sentence summaries $a(x)$ for $x \in g$.  
    Each summary is embedded via a text encoder $v$, yielding vectors $v(a(x)) \in \mathbb{R}^{d'}$.  
    We then identify a subset
    \begin{equation}
        \mathcal{B}_t \subset S, \quad |\mathcal{B}_t| \geq 5
    \end{equation}
    such that the embeddings $\{g(a(x)) : x \in \mathcal{B}_t\}$ form a tight cluster in embedding space (e.g., via average pairwise similarity).  
    This ensures that each deliberation batch centers on a \emph{single coherent ambiguity dimension}.

\end{enumerate}

\vspace{1ex}
\noindent\textbf{Concept refinement.}
In each iteration, users produce labels $y(x) \in \{0,1\}$ for $x \in \mathcal{B}_t$ and optional textual comments $c(x)$.
We refine the concept definition via automatic prompt optimization, following prior work~\cite{Pryzant2023AutomaticPO, prasad2023grips}.
User comments are expanded into explicit rationales $r_{\text{user}}(x)$ (e.g., turning brief notes into full sentences).
An LLM then synthesizes a set of $M$ candidate definitions
\begin{equation}
    \mathcal{C}_t = \bigl\{ d_t^{(1)}, \dots, d_t^{(M)} \bigr\}
\end{equation}
each of which modifies $d_t$ to reflect the new rationales in different ways (e.g., tightening constraints on creamy sauces, clarifying thresholds on sugar content). Given the current labeled set $\mathcal{L}_{t}$, we evaluate each candidate by prompting the VLM with $d_t^{(m)}$ and computing $F_1$ against user labels across all rounds so far. We choose the best candidate via a greedy update:
\begin{equation}
    d_{t+1} = \arg\max_{m} F_1\bigl(f_{d_t^{(m)}},\, \mathcal{L}_{t}\bigr)
\end{equation}
Greedy selection across rounds makes the evolution of $d_t$ transparent to users, in contrast to more complex search procedures such as beam search~\cite{Pryzant2023AutomaticPO} or Markov-chain-based exploration~\cite{deng2022rlprompt}, which would be harder to inspect~\cite{spinner2025revealing} and would demand substantially more computation—both undesirable in a real-time deliberation loop.


\subsection{Implementation details}
We use Gemini-Pro 2.5 for the concept decomposition component and Gemini-Flash 2.5 for all other tasks in our implementation, including image classification, borderline query generation, and ambiguity mining~\cite{team2023gemini, comanici2025gemini}.
The choice of foundation models is based on their state-of-the-art performance at the time of writing.
These models are publicly available through the Google Cloud API~\cite{google2025gemini}, and have not been further trained or fine-tuned in this work—an intentional decision to ensure accessibility for domain experts regardless of compute resources.
For image retrieval, we employ an existing nearest-neighbors search implementation~\cite{guo2016quantization} to return visually similar images for a given text query from a large unlabeled dataset.
The user interface is implemented in Google Colab~\cite{googlecolab}; all user interactions and model calls can be invoked through an interactive notebook.
See Appendix \ref{parameter_appendix} for detailed parameter values.

%% file: 04_experiments.tex
\section{Experiments}

Evaluating subjective visual classification is inherently challenging as it lacks the static, objective ground truth of standard benchmarks. Because user definitions are fluid and evolve during deliberation, fixed datasets and LLM simulations cannot capture the dynamic alignment between a user’s shifting mental model and the classifier. We therefore evaluated Agile Deliberation through \textbf{18 live user sessions, each $1.5h$ long}. This setup allows us to measure the system's success in articulating unique human intent—a critical metric that offline evaluation cannot assess.

Results show that:
(1) Agile Deliberation consistently outperforms both automated baselines and manual deliberation, achieving up to an 11\% improvement in $F_1$ over zero-shot classifiers;
(2) these gains stem from clearer concept articulation and better human–AI alignment, as participants explored diverse borderline cases and offloaded prompt optimization to the system; and
(3) participants reported significantly lower workload and frustration, unanimously preferring Agile Deliberation for its structured workflow.

\vspace{1ex}
\noindent\textbf{Baselines.}
We compared our approach against three baselines: two fully automated systems without concept deliberation and one human-in-the-loop condition.
\begin{itemize}
    \item The first automated baseline, \textit{Zero-Shot Learning}, classifies each image directly using the initial concept description using Gemini-Flash 2.5.
    \item The second, \textit{Modeling Collaborator}~\cite{toubal2024modeling}, automatically decomposes the concept into detailed prompts via prompt chaining (using Gemini-Pro 2.5 for fair comparison) and uses these for image classification.
    \item In \textit{Manual Deliberation}, participants manually explore borderline images in the target dataset using an image search engine and iteratively refine their prompts accordingly. To make this baseline even stronger, participants were also given access to the detailed prompts generated by \textit{Modeling Collaborator} as optional references.
\end{itemize}

\vspace{1ex}
\noindent\textbf{Participants \& recruitment.}
Both Agile and Manual Deliberation required user studies for evaluation.
We recruited nine participants \textit{without prior professional content moderation experience} through an open call within our organization, representing diverse demographics.
Each participant completed two sessions on different concepts, one using Agile Deliberation and another Manual Deliberation, totaling $18$ sessions.
Each session lasted approximately $90$ minutes, and participants received $\$100$ as compensation.
The substantial time required for these studies limited the total number of sessions.
System order was randomized across participants to mitigate potential order effects.

\vspace{1ex}
\noindent\textbf{Experiment procedures.}
In both systems, participants first completed an onboarding session covering the study overview and system tutorial.
Starting with only a concept name and short description, they were encouraged to develop their own understanding of the concept.
They then spent about 45 minutes articulating a concept definition that could directly serve as a performant image classifier.
Afterward, participants annotated 200 images from a held-out test set to evaluate classifier performance and completed a short survey on their experience.
Each session concluded with a 15-minute semi-structured interview to gather qualitative insights into their reasoning and overall experience.

Ideally, participants would have used both systems to deliberate on the same concept, enabling direct comparison of classifier performance and subjective experience (e.g., usability and interaction flow) within individuals.
However, this design was infeasible: once a participant defined a concept in the first session, they had already established a clear conceptual boundary, making a second round on the same concept highly biased.
We therefore assigned different concepts to the two systems.
This allows comparing subjective experience \textit{within participants}, but classifier performance only \textit{between participants}, since each system was trained on a different concept.
Because participants varied in their deliberation ability and concept complexity, such between-participant comparisons are only meaningful with enough participants per concept.
Given limited study sessions, we prioritized assigning more participants per concept over testing more concepts with fewer participants each.

\vspace{1ex}
\noindent\textbf{Concepts and data sources.}
We selected two concepts for our study: (1) \texttt{paid to play}, a real-world moderation concept involving images that promise unrealistic rewards for online entertainment as clickbait, and (2) \texttt{healthy food}.
These concepts were chosen because they are shown to be highly subjective and semantically complex, making them well-suited for evaluating systems that support concept deliberation~\cite{stretcu2023agile}.
They also represent two application contexts for image classifiers: \texttt{paid to play} captures platform-level moderation mitigating harmful content, while \texttt{healthy food} captures end-user curation promoting desirable content.
We followed the approach of Agile Modeling~\cite{stretcu2023agile} to sample $200$ images per concept as the test datasets, covering the full spectrum of examples within each domain.
For the target dataset, we used a proprietary domain-specific $100M$ image dataset for \texttt{paid to play} and the WebLI dataset~\cite{chen2022pali} for \texttt{healthy food}.


\input{table_classifer-results-boxes-horizontal}

\subsection{More Gains in Classification Performances}

We show our experiment results in Table \ref{tab:performance}.
We measure the alignment between participants and their image classifiers using precision, recall, and $F_1$ scores.
We did not use accuracy because participants’ personalized concept definitions led to imbalanced label distributions in the test set.

\vspace{0.7ex}
\textbf{Participants using Agile Deliberation achieved higher performance than two automated baselines.}
Their zero-shot classifiers achieved $F_1$ scores below $0.5$, indicating that both concepts required nuanced interpretations beyond general VLM priors.
Compared to two automated baselines, participants using Agile Deliberation achieved substantially higher $F_1$ scores—$10.5\%$ and $7\%$ on average above zero-shot classifiers and Modeling Collaborator, respectively.
These gains were mainly driven by improved precision with only minor reductions in recall, reflecting how zero-shot classifiers often make overly broad predictions~\cite{toubal2024modeling}.
In contrast, Modeling Collaborator, where LLMs automatically enriched the concept definition without any user feedback, led to only modest improvements over the zero-shot baseline, suggesting that automated refinement alone cannot capture users’ nuanced intentions. 
Overall, these results show that iterative human feedback in Agile Deliberation enables finer alignment between human concept understanding and classifier decisions.

For participants using Manual Deliberation, a similar trend emerged for the concept \texttt{paid to play}, underscoring the value of concept deliberation.
However, for \texttt{healthy food}, the trend reversed: both Modeling Collaborator and Manual Deliberation yielded slightly lower $F_1$ scores than the zero-shot baseline.
Since their zero-shot classifiers already performed well, participants using Manual Deliberation likely had an easy concept understanding, leaving limited room for further deliberation.
These findings indicate that \textbf{concept deliberation offers greater value when users hold more nuanced interpretations that deviate from generic VLM priors.}

\begin{figure}
    \centering
    \includegraphics[width=\linewidth]{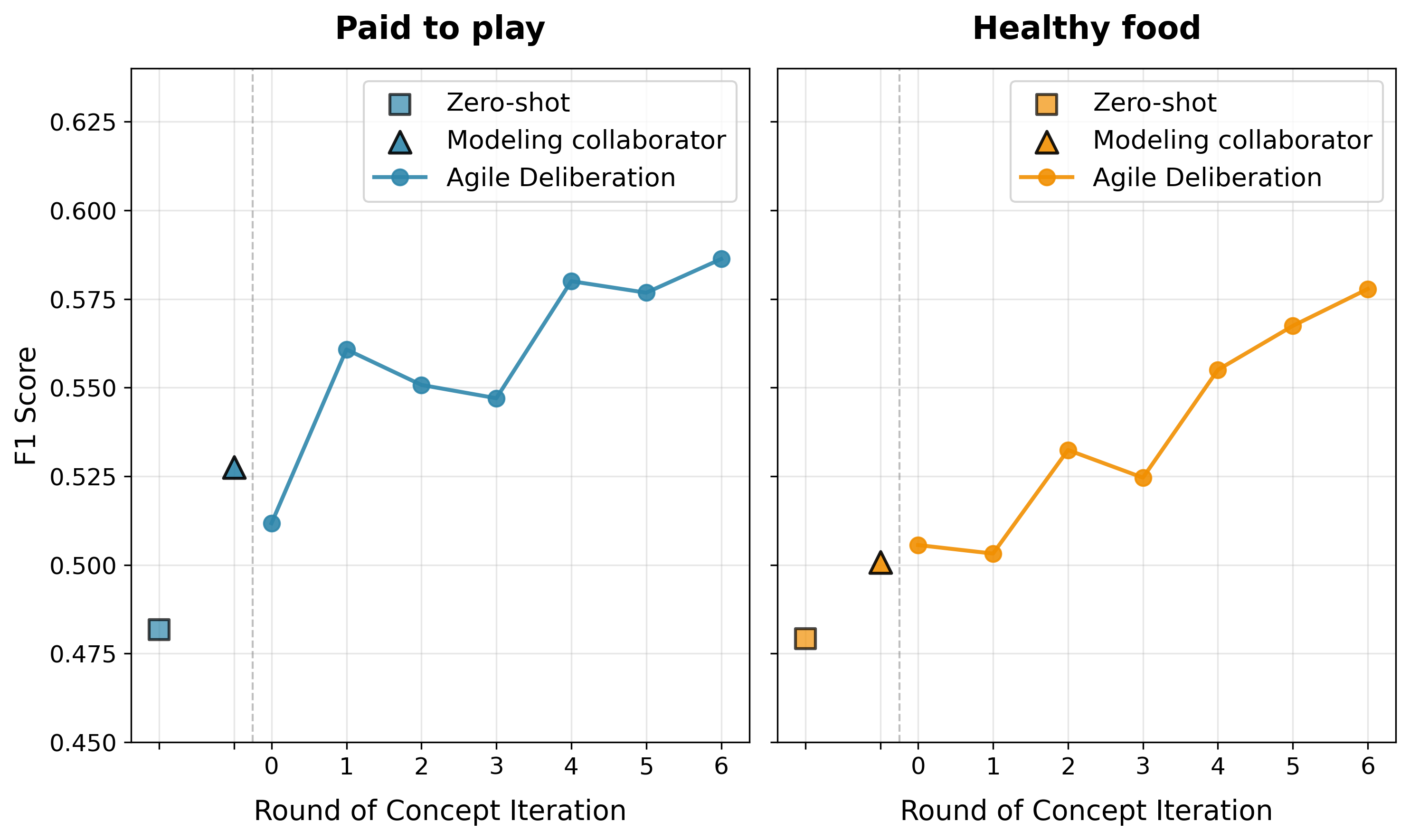}
    \vspace{-2.5ex}
    \caption{$F_1$ scores of Agile Deliberation across rounds of concept iteration compared with two automated baselines.}
    \vspace{-4ex}
    \hfill
    \label{agile_modeling_across_rounds}
\end{figure}

\vspace{0.7ex}
\textbf{Participants using Agile Deliberation achieved higher performance gains than those using Manual Deliberation.}
Because participants varied in concept complexity, we compared each system’s \emph{relative improvement} over the zero-shot baseline rather than absolute scores.
For \texttt{Paid to Play}, Agile Deliberation improved $F_1$ scores by $11\%$, compared to an $8\%$ gain with Manual Deliberation.
For \texttt{Healthy Food}, Agile Deliberation yielded an $10\%$ improvement, whereas Manual Deliberation showed a slight decrease ($-3\%$).
Together, these results suggest that Agile Deliberation delivers consistent improvements in classifier performance for subjective concepts.

Figure~\ref{agile_modeling_across_rounds} further shows how the $F_1$ scores of Agile Deliberation participants evolved across rounds of concept iteration.
Performance fluctuated but trended upward for both concepts.
Although the number of rounds was limited by session duration, this trend suggests that longer deliberation could yield further gains.
The fluctuations partly reflect the system’s real-time design—our prompt optimization algorithm prioritized interactive responsiveness under limited computational resources rather than exhaustive refinement.

\subsection{Lower Barriers for Concept Deliberation.}

\input{table_survey-boxes}

\textbf{Performance gains of Agile Deliberation resulted from both clearer concept understanding and better human-VLM alignment.}
Participants reported in the final interviews that Agile Deliberation made it easier to identify borderline cases for their subjective concepts than Manual Deliberation.
Participants using Manual Deliberation tried an average of 7.3 search queries when finding borderline images, often focusing on a narrow type of ambiguity.
For example, one participant repeatedly searched for images of healthy food overlapped with fried food (e.g., \textit{vegetable smoothie with fried chicken}, \textit{fried fish with vegetables}).
P7 summarized this frustration: ``\textit{I think it's hard for me to think of queries that will contradict my definitions because sometimes I don't know what is the borderline for my definitions.}''
By contrast, participants using Agile Deliberation explored more diverse ambiguities, including food not being the main subject, meals with high-carbohydrate ingredients, or dishes still under preparation.
Participants also reported that refining prompts in Manual Deliberation was difficult, echoing prior observations in end-user prompt engineering~\cite{wang2025end, zamfirescu2023johnny}.
Some worried that their edits ``\textit{might overfit, making things worse in other images.}''
Agile Deliberation mitigated these issues by automating prompt optimization, allowing participants to provide feedback in casual natural language.
As one participant explained, ``\textit{This is more convenient because LLMs take in feedback from people, summarize it, and put back its revision.}''

\vspace{0.5ex}
\textbf{Participants found it easier to deliberate with Agile Deliberation than with Manual Deliberation, unanimously preferring the former.}
Survey results (Table~\ref{user-experience}) show that participants exerted significantly less effort to achieve good performance in the Agile condition than in the Manual condition ($M$ = 3.11 vs 4.67, $p < .05$).
They also reported fewer negative emotions such as stress or irritation ($M$ = 1.67 vs. 3.00; $p < .05$).
All participants preferred Agile Deliberation over Manual Deliberation.
One participant noted, ``\textit{[Agile Deliberation] made less toil on me because it listed out edge cases in a comparison,}'' while another described it as ``\textit{a more structured and systematic approach.}''
Together, these findings suggest that \textit{Agile Deliberation} lowers the barrier for non-experts to translate their subjective concepts into visual classifiers.

%% file: table_classifer-results-boxes-horizontal.tex
\begin{table*}[t!]
\vspace{-3ex}
\centering
\footnotesize
\lato
\def\arraystretch{1.3}
\setlength{\tabcolsep}{1.5pt}
\begin{tabularx}{\linewidth}{|l|c||Y|Y|Y||Y|Y|Y|}
    \hhline{~-||---||---|}
        \rowcolor{gray!15}
        \multicolumn{1}{c|}{\cellcolor{white}} &  &  \multicolumn{3}{c||}{Participants in \textbf{Agile Deliberation Condition}} &  \multicolumn{3}{c|}{Participants in \textbf{Manual Condition Condition}} \\
        \hhline{|~~||---||---|}
        \rowcolor{gray!15}
        \multicolumn{1}{l|}{\cellcolor{white}} & \multirow{-2}{*}{Condition} & F1 & Precision & Recall &  F1 & Precision & Recall \\
    \hhline{-=::===::===}
        \cellcolor{gray!15} & Zero-shot & 0.48 (.07) & 0.35 (.07) & \textbf{0.80} (.12)  & 0.43 (.08) & 0.29 (.06) & \textbf{0.84} (.12)\\
        \cellcolor{gray!15} & Modeling Collaborator & 0.53 (.09) & 0.47 (.12) & 0.63 (.14) & 0.47 (.14) & 0.40 (.14) & 0.58 (.16)\\
        \cellcolor{gray!15} & Assigned Deliberation System & \textbf{0.59} (.06) & \textbf{0.57} (.13) & 0.63 (.11)  & \textbf{0.51} (.05) & \textbf{0.42} (.07) & 0.68 (.07) \\
        \cellcolor{gray!15} \multirow{-4}{*}{\rotatebox[origin=c]{90}{\textbf{Paid to play}}} & $\Delta$ & 11\% & 22\% & -17\% & 8\% & 13\% & -16\% \\[0pt]
    \hhline{:=:=::===::===:}
        \cellcolor{gray!15} & Zero-shot & 0.48 (.15) & 0.33 (.13) & \textbf{0.99} (.02) & \textbf{0.82} (.02) & 0.73 (.05) & \textbf{0.92} (.03) \\
        \cellcolor{gray!15} & Modeling Collaborator & 0.50 (.14) & 0.36 (.15) & 0.91 (.06)  & 0.78 (.03) & 0.77 (.06) & 0.79 (.02) \\
        \cellcolor{gray!15} & Assigned Deliberation System & \textbf{0.58} (.10) & \textbf{0.44} (.10) & 0.84 (.07)& 0.79 (.07) & \textbf{0.80} (.03) & 0.79 (.13)  \\
        \cellcolor{gray!15} \multirow{-4}{*}{\rotatebox[origin=c]{90}{\textbf{Healthy food}}} & $\Delta$ & 10\% & 11\% & -15\% & -3\% & 7\% & -13\% \\[0pt]
    \hhline{|-|-||---||---|}
\end{tabularx}

\vspace{-1ex}
\caption{
    \textbf{Classification performance across participant conditions for both concepts.} $F_1$, precision, and recall are averaged over participants in each condition (standard deviation in parentheses). ``Assigned Deliberation System'' denotes Agile or Manual per condition. Since participants varied in the complexity of their understanding for the same concept, the classification performances are not comparable across these two groups. Instead, we focus on each system’s relative improvement over its respective zero-shot baseline (indicated by $\Delta$).}
\label{tab:performance}
\vspace{-3ex}
\end{table*}

%% file: table_survey-boxes.tex
\begin{table}[t!]
\centering
\vspace{1ex}
\footnotesize
\lato
\def\arraystretch{1.3}
\begin{tabularx}{\linewidth}{|L|c|c|}
    \hhline{|---|}
        \rowcolor{gray!15}
         & Agile (ours) & Manual \\
        \rowcolor{gray!15}
         \multicolumn{1}{|>{\centering\arraybackslash}X|}{\multirow{-2}{*}{Survey Item}} & M (SD) & M (SD) \\
    \hhline{=:=:=}
  Effort required to achieve good performance ($\downarrow$ is better) & 3.11 (1.62)* & 4.67 (0.71)* \\
  \hhline{|---|}

    Success in articulating concept ideas ($\uparrow$ is better) &  5.56 (0.88) & 5.11 (1.54)  \\
   \hhline{|---|}

    Frustration while defining the concept ($\downarrow$ is better) &  1.78 (1.09) & 2.33 (0.87) \\
   \hhline{|---|}

    Feeling insecure, stressed, or annoyed ($\downarrow$ is better) & 1.67 (0.71)* & 3.00 (1.41)* \\
   \hhline{|---|}

    Ease of creating a comprehensive definition ($\uparrow$ is better) & 4.11 (1.27) & 4.00 (0.50) \\
   \hhline{|---|}

    Mental demand when defining the concept ($\downarrow$ is better) & 3.22 (2.05) & 4.56 (1.01) \\
   \hhline{|---|}
\end{tabularx}

\vspace{-1ex}
\caption{Comparison of user ratings between Agile and Manual Deliberation. $M$ = Mean, SD = Standard Deviation. * $p < .05$. }
\label{user-experience}
\vspace{-4ex}
\end{table}

%% file: 06_conclusion.tex
\section{Conclusion}

We introduced \textit{Agile Deliberation}, a human-in-the-loop framework for classifying subjective and evolving visual concepts. By guiding users through structured concept scoping and surfacing borderline examples for reflection, our system helps users clarify their ambiguous mental models. Our user study—prioritizing live interaction over static benchmarks to capture genuine concept evolution—demonstrated that this approach not only yields more performant classifiers than manual or fully automated baselines but also significantly reduces the cognitive burden associated with defining complex concepts. As visual classification increasingly moves into subjective domains, frameworks like Agile Deliberation will be essential for aligning powerful AI models with nuanced human intent. Limitations and future work are discussed in the appendix.~\ref{limitations_appendix}.

%% file: appendix.tex
\clearpage
\appendix

\section{Limitations}
\label{limitations_appendix}

Here we discuss a few limitations of our study and how these can be addressed in future work. Most of these are caused by the extensive costs required to conduct more user studies, as explained below.

\vspace{1ex}
\noindent\textbf{User study scale.}
Our evaluation involved nine participants across $18$ sessions (approximately $27$ hours of study time), which may limit the statistical power and generalizability of our findings. Each session required extended interaction with the system to iteratively refine subjective concept definitions, leading us to prioritize depth of interaction over study scale. The total study time is comparable to prior work at the intersection of HCI and ML systems evaluating subjective, contextual workflows, which often report 10–30 hours of user study time~\cite{wang2025promptimizer,lam2025policy}. Future work should examine larger participant pools and a broader set of concepts to strengthen external validity.

\vspace{1ex}
\noindent\textbf{Component-level ablations.}
Our evaluation does not isolate the contributions of individual components within Agile Deliberation, such as borderline image retrieval and prompt optimization. Instead, we evaluate the system as an integrated workflow. This design reflects how these mechanisms operate together in practice: formative interviews with domain experts suggested that identifying informative borderline examples and refining definitions are tightly coupled activities in concept deliberation. Future work should conduct additional user studies to independently examine the effects of these components.

\vspace{1ex}
\noindent\textbf{Evaluation across generative models.}
Our implementation of Agile Deliberation primarily relies on a single generative model (Gemini~2.5 Flash), and we did not evaluate the full pipeline across other generative models. On one hand, Agile Deliberation only requires inference access and can in principle be used with any API-accessible VLM, including newer models with stronger reasoning capabilities. Future work should evaluate the framework with a broader range of models to better understand how model choice affects system performance and usability.

Beyond the generalizability of the full pipeline, there is also the question of whether concept definitions produced by Agile Deliberation transfer effectively to other VLMs. We provide a preliminary investigation of this question in the next section. Future work should more systematically evaluate the transferability of concept definitions across a broader range of models.

\section{Transferability of Concept Definitions Across VLMs}

We conducted an additional experiment to examine whether concept definitions produced by Agile Deliberation transfer to other VLMs. 
Specifically, we took concept definitions produced under both the Agile Deliberation condition and the baseline condition—both implemented using \textit{Gemini~2.5 Flash}—and applied them to external, smaller VLMs for classification: \textit{Qwen3-VL-8B} and \textit{Qwen3-VL-30B-A3B-Instruct}~\cite{qwen3technicalreport}. We presented the results at the Table~\ref{tab:performance-generalization}.

Overall, the results suggest that concept definitions from both conditions transfer to other models, but Agile Deliberation produces definitions that are more sensitive to model capability. Definitions from the Manual condition remain relatively similar across models, while Agile Deliberation definitions continue to show benefits on the Qwen models, with the strongest gains appearing on more capable models. This pattern suggests that Agile Deliberation yields richer and more complex concept definitions, whose advantages become more visible as model reasoning ability improves.

It is also worth emphasizing that these improvements are obtained despite the fact that Agile Deliberation optimized the concept definition for a different model (i.e. \textit{Gemini~2.5 Flash}), and we expect even bigger gains when the target model is also the one used as classifier in the Agile Deliberation process.

\input{table_concept_generalization}

\section{Interview Analysis}
\label{interview_appendix}
\subsection{Concept Deliberation by Experts}
We first conducted a qualitative analysis of 20 concept definitions created by professional content moderators as part of their regular workflow. These materials included both the finalized definitions and their accompanying discussion threads, giving us visibility into how teams deliberate over subjective visual concepts. 
We learned that practitioners typically begin by \textit{scoping} their concept definitions and then iteratively \textit{refining} them by searching for and reflecting on borderline images.
Both of these stages require domain expertise and newcomers in the team need to spend significant time familiarizing themselves with this process.

\vspace{1ex}
\textbf{Stage 1: Concept Scoping — Structured definitions facilitate concept deliberation.}
Experts often structured their initial definitions by decomposing composite concepts into simpler unit components. For example, a composite idea such as \textit{before-and-after transformations for achievements} was separated into the visual pattern of a \textit{before–after layout} and the semantic notion of an \textit{achievement}, whereas a concept like \textit{beautiful images} was treated as a single unit. Each unit concept was then expanded into positive and negative visual categories—for instance, \textit{city parks} versus \textit{industrial factories} when reasoning about \textit{beautiful images}. This hierarchical and contrastive structure helped practitioners articulate core visual signals and establish the initial decision boundary.

\vspace{1ex}
\textbf{Stage 2: Concept Iteration — Borderline images are essential for iterative refinement.}
Borderline images then played a critical role in refining these scoped definitions. Such images revealed ambiguities that the initial structure did not address—for example, encountering an industrial site that had been remodeled into a park prompted practitioners to reconsider how transitional or mixed scenes should be classified under \textit{beautiful images}. These borderline cases made gaps in the definition immediately visible and helped practitioners clarify ambiguous language, resolve inconsistencies, and sharpen the intended scope. In practice, visual inspection of borderline images proved far more intuitive than abstract reasoning alone, providing concrete signals that guided iterative refinement.

\subsection{Challenges in Concept Deliberation}
Our analysis of 20 expert-authored concept definitions gave us an initial picture of how subjective concepts are scoped and refined. Building on this insight, we conducted five formative studies with professional content moderators to better understand the challenges they encounter.
Each session was one hour long and consisted of two parts.
In the first part, participants defined a new visual concept using a provided template. They scoped the concept definition, searched for borderline images in a think-aloud manner. In the second half, we conducted a semi-structured interview to discuss the challenges they encountered, how they searched for borderline images, and how they expressed nuanced decision boundaries when preparing a concept for an LLM-based classifier.

Even with substantial expertise, participants still reported several difficulties. They spent a large amount of time searching for edge-case images that could clarify the subtle boundaries of subjective concepts. Although initial scoping was usually straightforward, identifying truly borderline images was much harder. It required them to question their own intuitions, and locating such cases in a large dataset was slow and often frustrating. Existing tools provided only limited support. Similarity search tended to surface images that were close in embedding space but not genuinely borderline, so these results offered little insight into conceptual edges. Participants also tried using LLMs to generate search queries, but they found that LLMs often lacked awareness of the visual complexity in real datasets. Many borderline cases involved cropped, blurred, zoomed-in, or partially occluded images, or images embedded in cluttered backgrounds. These subtleties were difficult for LLMs to reason about~\cite{luo2023knowledge, wu2025visualized}. 
As a result, experts still relied heavily on manual exploration to identify the borderline cases that were most helpful for refining their definitions.

\section{Interface of Agile Deliberation}

\begin{figure*}
    \centering
    \includegraphics[width=0.90\textwidth]{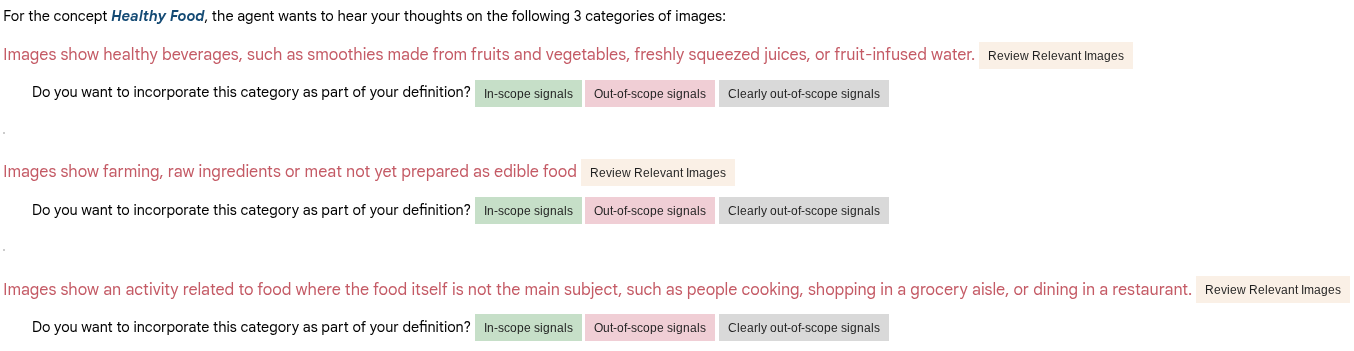}
   \caption{\textbf{Concept Scoping interface.} The system proposes candidate positive and negative subconcepts derived from the user's initial concept description. For each subconcept, representative images retrieved from the dataset are shown for inspection. The user decides whether the subconcept should be included in the structured concept definition.}
    \hfill
    \label{concept_scoping_round}
    \vspace{-1em}
\end{figure*}

\begin{figure*}
    \centering
    \includegraphics[width=0.90\textwidth]{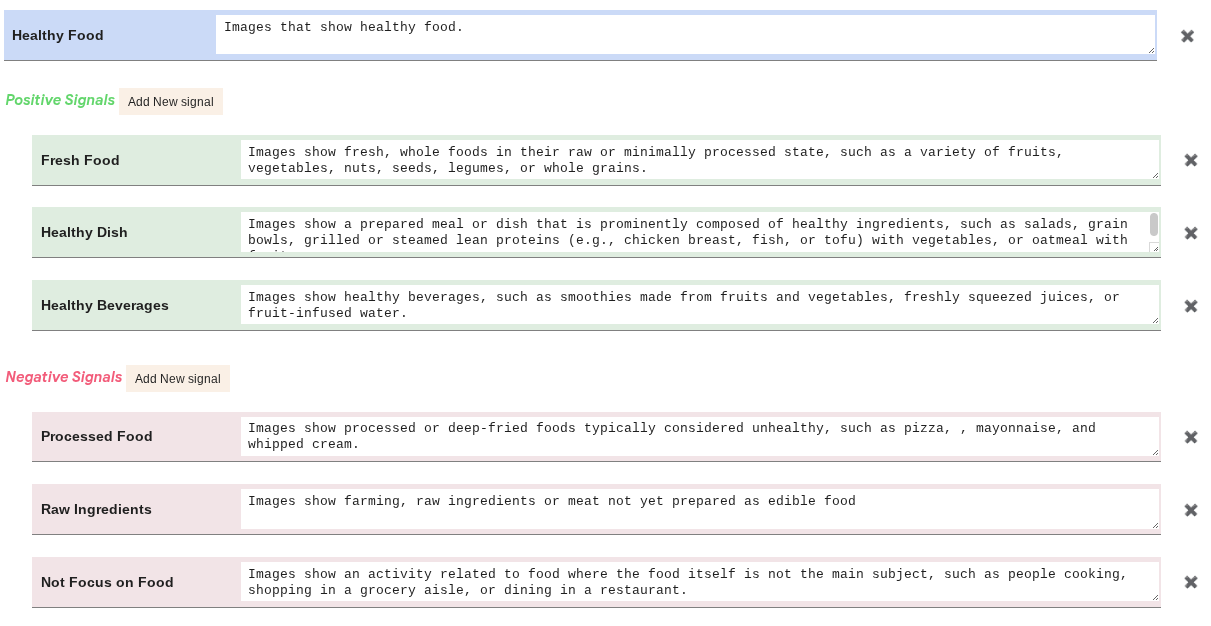}
    \caption{\textbf{Structured concept definition after Concept Scoping.} Based on the user's selections, the system constructs an initial concept definition composed of positive and negative subconcepts that capture the core visual signals of the concept.}
    \hfill
    \label{concept_scoping_result}
    \vspace{-1em}
\end{figure*}

\begin{figure*}
    \centering
    \includegraphics[width=0.90\textwidth]{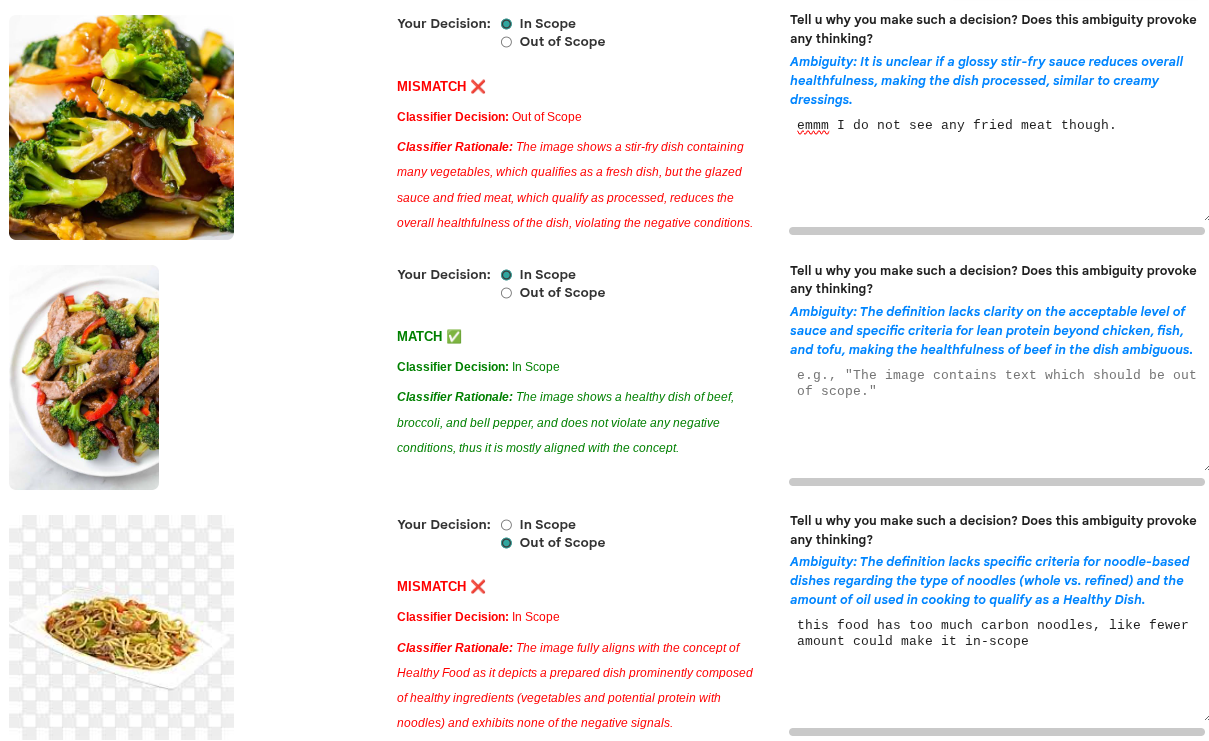}
    \caption{\textbf{Concept Iteration interface.} The system retrieves a batch of borderline images that are semantically ambiguous under the current concept definition. The user reviews the model's prediction and explanation and then labels each image as in-scope or out-of-scope.}
    \hfill
    \label{concept_iteration_round}
    \vspace{-1em}
\end{figure*}

\begin{figure*}
    \centering
    \includegraphics[width=0.90\textwidth]{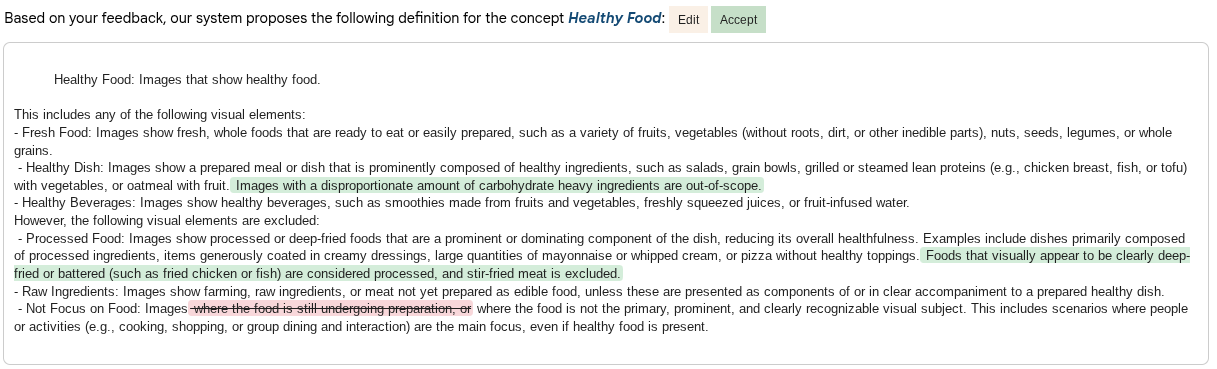}
    \caption{\textbf{Updated concept definition after one iteration round.} User feedback on borderline images is incorporated to refine the concept definition, producing an improved classifier aligned with the user's interpretation.}
    \hfill
    \label{concept_iteration_result}
    \vspace{-1em}
\end{figure*}

Figure~\ref{concept_scoping_round}--\ref{concept_iteration_result} show screenshots from the interactive notebook interface used in our study. The interface supports the two stages of Agile Deliberation: \textit{concept scoping}, where users review candidate subconcepts derived from an initial concept description, and \textit{concept iteration}, where users inspect and label borderline images to refine the concept definition over multiple rounds.

\section{LLM Prompts}
\label{prompts_appendix}
\subsection{Image Classifiers}
\textbf{Prompts that classify images based on the structured definition.} We convert the resulting ratings into binary labels by thresholding at 3.
\input{prompt_simple_concept_evaluate_image_prompt}

\subsection{Decomposition module}
\textbf{Prompts that decompose a composite concept}
\input{prompt_decompose_concept_prompt}

\textbf{Prompts that brainstorm candidate categories of a given concept}
\input{prompt_brainstorm_golden_category_prompt}

\textbf{Prompts that brainstorm more borderline categories of a given concept}
\input{prompt_brainstorm_borderline_category_prompt}

\subsection{Borderline Image Retrieval Module}
\textbf{Prompts that generate borderline/in-scope descriptions of the subjective concept}
\input{prompt_generate_descriptions_prompt}

\textbf{Prompts that examine ambiguities of individual images}

\input{prompt_whether_borderline_prompt}

\subsection{Concept refinement}

\textbf{Prompts that articulate user feedback into explicit rationales}

\input{prompt_reflect_classified_image_prompt}

\newpage
\textbf{Prompts that generate refinement candidates}
\input{prompt_improve_definition_prompt}

\section{System Parameters}
\label{parameter_appendix}
\noindent\textbf{Borderline Image Retrieval Module.}
To select which cluster to query next, we compute an interaction-based score for every cluster using four heuristics derived from users' previous interactions. For each cluster, we track how many images have been explored, how often users marked them as mistakes, how often users provided textual feedback, and the distribution of user-provided ratings. We then compute:

\begin{itemize}[leftmargin=12pt]
    \item \textbf{Mistake rate.} The proportion of explored images that users labeled as incorrect for the current definition. Clusters with higher mistake rates are prioritized because they reveal boundary mismatches.
    
    \item \textbf{Feedback rate.} The proportion of explored images for which users provided any written justification. Clusters that elicit richer feedback are treated as more informative.
    
    \item \textbf{Exploration value.} A term that favors clusters with many unseen images. This prevents repeatedly sampling from clusters that are already heavily explored.
    
    \item \textbf{Diversity rate.} The standard deviation of user-provided ratings within the cluster. Higher diversity indicates conceptual ambiguity and thus higher expected value for refinement.
\end{itemize}

These four components are combined into a weighted score:
\begin{equation}
\begin{aligned}
\text{Score} =\;& 0.5\,\text{mistake\_rate} + 0.3\,\text{feedback\_rate} \\
              & + 0.15\,\text{exploration\_value} \\& +  0.05\,\text{diversity\_rate}.
\end{aligned}
\end{equation}
where the coefficients were chosen based on preliminary experiments balancing exploration and refinement. The cluster with the highest score is selected for the next round of iteration.

\vspace{1ex}
\noindent\textbf{Ambiguity mining}.
To surface a coherent batch of $5$ borderline images per round, we sample $25$ candidates, generate short ambiguity summaries, embed them, and cluster the embeddings to isolate a single ambiguity dimension. We use an adaptive DBSCAN radius to identify a cluster of at least $5$ images. The full clustering procedure is provided below.

\begin{algorithm}[h]
\caption{Adaptive DBSCAN for Ambiguity Clustering}
\begin{algorithmic}[1]
\Require embeddings $\{e_i\}$;
         minimum cluster size $k=5$;
         minimum radius $\varepsilon_{\min}=0.2$;
         maximum radius $\varepsilon_{\max}=0.8$;
         step size $\Delta\varepsilon = 0.01$
\Ensure a cluster of size $\ge k$
\State $\varepsilon \gets \varepsilon_{\min}$
\While{$\varepsilon \le \varepsilon_{\max}$}
    \State labels $\gets$ \textsc{DBSCAN}$(\varepsilon)$ on $\{e_i\}$
    \State clusters $\gets$ non-noise groups in labels
    \If{any cluster has size $\ge k$}
         \State \Return that cluster
    \EndIf
    \State $\varepsilon \gets \varepsilon + \Delta\varepsilon$
\EndWhile
\State \Return clusters from final $\varepsilon$
\end{algorithmic}
\end{algorithm}

\vspace{1ex}
\noindent\textbf{Concept refinement}.
Based on the user’s feedback for the current batch, the system generates $5$ candidate refinements $\{d_t^{(1)}, \dots, d_t^{(5)}\}$ of the current definition $d_t$. We first filter the top $3$ candidates based on their performance on the current borderline set $\mathcal{B}_t$, ensuring that the new rationales are directly reflected. Among these $3$ candidates, we then select the final update $d_{t+1}$ according to their performance on the full labeled set $\mathcal{L}_t$ accumulated so far. This two-stage procedure incorporates user feedback immediately while preserving global consistency across rounds.
      
\section{Example Definitions}
\label{definitions_appendix}

Here is one full example of the definition $d_0$ after the scoping stage.

\begin{tcolorbox}[
    breakable,
    colback=white,
    colframe=black!25,
    boxsep=1mm,
    left=1mm,right=1mm,top=1mm,bottom=1mm,
]
\vspace{1ex}
\noindent\textbf{This includes any of the following visual elements:}
\begin{itemize}
    \item \textit{Healthy Dish}: Images show a prepared meal or dish that is prominently composed of healthy ingredients, such as salads, grain bowls, grilled or steamed lean proteins (e.g., chicken breast, fish, or tofu) with vegetables, or oatmeal with fruit.
    \item \textit{Healthy Beverages}: Images show healthy beverages, such as smoothies made from fruits and vegetables, freshly squeezed juices, or fruit-infused water.
\end{itemize}

\noindent\textbf{However, the following visual elements are excluded:}
\begin{itemize}
    \item \textit{Processed Food}: Images show processed foods typically considered unhealthy, such as pizza, mayonnaise, or whipped cream.
    \item \textit{Raw Ingredients}: Images show farming scenes, raw ingredients, or meat that is not yet prepared as edible food.
    \item \textit{Not Focus on Food}: Images show an activity related to food where the food itself is not the main subject, such as people cooking, shopping in a grocery aisle, or dining in a restaurant.
\end{itemize}
\end{tcolorbox}

%% file: table_concept_generalization.tex
\begin{table*}[t!]
\centering
\footnotesize
\lato
\def\arraystretch{2}
\setlength{\tabcolsep}{1.5pt}
\begin{tabularx}{\linewidth}{|l|c||Y|Y|Y||Y|Y|Y|}
    \hhline{~-||---||---|}
        \rowcolor{gray!15}
        \multicolumn{1}{c|}{\cellcolor{white}} & Model:\textbf{Qwen3-VL-8B}  &  \multicolumn{3}{c||}{Participants in \textbf{Agile Deliberation Condition}} &  \multicolumn{3}{c|}{Participants in \textbf{Manual Condition Condition}} \\
        \hhline{|~-||---||---|}
        \rowcolor{gray!15}
        \multicolumn{1}{l|}{\cellcolor{white}} & \multirow{-1}{*}{Condition } & F1 & Precision & Recall &  F1 & Precision & Recall \\
    \hhline{-=::===::===}
        \cellcolor{gray!15} & Zero-shot & 0.46 (.09) & 0.31 (.08) & \textbf{0.95} (.07) & 0.36 (.06) & 0.22 (.05) & \textbf{0.93} (.06) \\ 
        \cellcolor{gray!15} & Assigned Deliberation System & \textbf{0.52} (.07) & \textbf{0.44} (.10) & 0.68 (.15) & \textbf{0.51} (.06) & \textbf{0.39} (.08) & 0.76 (.13)  \\
        \cellcolor{gray!15} \multirow{-3}{*}{\rotatebox[origin=c]{90}{\textbf{Paid to play}}} & $\Delta$ & 6\% & 13\% & -27\% & 15\% & 17\% & -17\% \\[0pt]
    \hhline{:=:=::===::===:}
        \cellcolor{gray!15} & Zero-shot & 0.47 (.14) & 0.32 (.13) & \textbf{0.96} (.03) & \textbf{0.81} (.04) & 0.73 (.06) & \textbf{0.91} (.02) \\
        \cellcolor{gray!15} & Assigned Deliberation System & \textbf{0.52} (.15) & \textbf{0.38} (.14) & 0.91 (.08) & \textbf{0.81} (.03) & \textbf{0.80} (.04) & 0.83 (.03) \\
        \cellcolor{gray!15} \multirow{-3}{*}{\rotatebox[origin=c]{90}{\textbf{Healthy food}}} & $\Delta$ & 5\% & 6\% & -5\% & 0\% & 7\% & -8\% \\[0pt]
    \hhline{|-|-||---||---|}
\end{tabularx}
\vspace{3ex}

\begin{tabularx}{\linewidth}{|l|c||Y|Y|Y||Y|Y|Y|}
    \hhline{~-||---||---|}
        \rowcolor{gray!15}
        \multicolumn{1}{c|}{\cellcolor{white}} & Model:\textbf{Qwen3-VL-30B-A3B-Instruct}  &  \multicolumn{3}{c||}{Participants in \textbf{Agile Deliberation Condition}} &  \multicolumn{3}{c|}{Participants in \textbf{Manual Condition Condition}} \\
        \hhline{|~-||---||---|}
        \rowcolor{gray!15}
        \multicolumn{1}{l|}{\cellcolor{white}} & \multirow{-1}{*}{Condition } & F1 & Precision & Recall &  F1 & Precision & Recall \\
    \hhline{-=::===::===}
        \cellcolor{gray!15} & Zero-shot & 0.49 (.09) & 0.34 (.08) & \textbf{0.97} (.05) & 0.38 (.09) & 0.24 (.06) & \textbf{0.93} (.08) \\
        \cellcolor{gray!15} & Assigned Deliberation System & \textbf{0.59} (.09) & \textbf{0.50} (.12) & 0.75 (.12) & \textbf{0.50} (.09) & \textbf{0.39} (.13) & 0.78 (.13) \\
        \cellcolor{gray!15} \multirow{-3}{*}{\rotatebox[origin=c]{90}{\textbf{Paid to play}}} & $\Delta$ & 10\% & 16\% & -22\% & 12\% & 15\% & -15\% \\[0pt]
    \hhline{:=:=::===::===:}
        \cellcolor{gray!15} & Zero-shot & 0.47 (.14) & 0.32 (.13) & \textbf{0.95} (.03) & \textbf{0.83} (.03) & 0.75 (.05) & \textbf{0.88} (.10) \\
        \cellcolor{gray!15} & Assigned Deliberation System & \textbf{0.52} (.19) & \textbf{0.40} (.18) & 0.83 (.21) & 0.78 (.04) & \textbf{0.79} (.04) & 0.78 (.05) \\
        \cellcolor{gray!15} \multirow{-3}{*}{\rotatebox[origin=c]{90}{\textbf{Healthy food}}} & $\Delta$ & 5\% & 8\% & -12\% & -5\% & 4\% & -10\% \\[0pt]
    \hhline{|-|-||---||---|}
\end{tabularx}

\caption{
    \textbf{Transfer performance of concept definitions across participant conditions on additional VLMs.}
    $F_1$, precision, and recall are averaged over participants in each condition (standard deviation in parentheses). We apply concept definitions produced under the Agile Deliberation and Manual conditions using \textit{Gemini~2.5 Flash} to additional VLMs to examine how well these definitions transfer across models. ``Assigned Deliberation System'' denotes Agile or Manual per condition. Since participants varied in the complexity of their understanding for the same concept, classification performances are not directly comparable across these two groups. Instead, we focus on each system’s improvement over its respective zero-shot baseline (indicated by $\Delta$).}
\label{tab:performance-generalization}
\vspace{-3ex}
\end{table*}

%% file: prompt_simple_concept_evaluate_image_prompt.tex
\begin{promptbox}
<role>You are an expert image annotator.</role>
<input>You will be provided with a single image, an image caption and a concept definition.</input>

<task>
- Thoroughly examine the image.
- Carefully consider all details of the concept
- Determine if the image satisfies every aspect of the given concept.
</task>

<step1>
Explicitly write down the specific requirements this concept demands. Make sure your decomposition is equivalent to the original definition. There are two special cases that demand your attention:
- If the concept is defined by a list of necessary conditions, you should start by checking each condition explicitly. You should only give a high rating if all conditions are satisfied.
- If the concept is defined by a list of positive and negative conditions, you should start by checking each condition explicitly. You should give a high rating if any of the positive conditions are satisfied and none of the negative conditions are satisfied.
- A concept can be first defined by a list of necessary conditions, and then each necessary condition can be further defined by a list of positive and negative conditions. You should follow the same logic recursively to check if the image satisfies the condition.
</step1>

<step2>
Based on responses in the first step, for each condition, you should determine whether the image satisfies the condition.
</step2>

<step3>
You should then combine your responses for individual conditions to determine whether the image satisfies the overall concept or not.
Your reasoning should be based solely on the definition, the image, and the image caption. Do not include any information that is not provided (no hallucinations).
</step3>

<step4>
Based on your reasoning in the first step, determine whether the image completely fulfills every aspect of the concept.
You should rate how in-scope the image is on a 1-5 Likert Scale where
- Rate 5 if the image fully aligns with the concept
- Rate 4 if the image mostly aligns with the concept; the small problem is a result of small ambiguities in the definition or small visual complexities in the image
- Rate 3 if there are no strong evidence that indicate that the image violates the concept, but the supporting evidence is also not strong enough to support a rating of 4 or 5.
- Rate 2 if the image violates some parts of the concept but there are some elements that are relevant to the concept.
- Rate 1 if the image does not align with the concept description at all.
</step4>

<step5>
Based on your answer in previous steps, provide a one-sentence summary why you give this answer.
</step5>

Provide your answer in the following XML structure:
<requirements>Explicitly write down the specific requirements this concept demands.</requirements>
<condition-eval>Write down your evaluation for each condition at step2 here.</condition-eval>
<evaluation>You should differentiate between concept definitions that requires the satisfaction of all conditions and those that only require the satisfaction of one of the conditions. Describe your evaluation reasonings in the step4 here</evaluation>
<decision>Rate on a 1-5 Likert Scale where 5 means the image is fully in-scope and 1 means the image is fully out-of-scope.</decision>
<summary>Provide your summary in the step5 here</summary>
The output will be later wrapped in a <root> tag, so do not wrap the content above in any tag such as xml, or root in your output.
\end{promptbox}

%% file: prompt_decompose_concept_prompt.tex
\begin{promptbox}
<role>You are an expert linguist recognized internationally.</role>
<input>You will receive a visual concept name and a description.</input>
<task>There are human image annotators who need to determine if an image is in scope or out of scope of this visual concept. Your job is to break down this visual concept into at most two necessary conditions: the conjunction of these necessary conditions will be logically equivalent to the given visual concept. In other words, images that satisfy all these conditions will be exactly images that satisfy the visual concept. We call this process as decomposition. The final goal is to make it easier and more accurate for human raters to determine if an image is in-scope or out-of-scope for the given visual concept.
</task>

<step1>
Rewrite the description by removing words that are redundant, too specific, or actually not necessary for this overall concept. Other than removing words, you should keep most parts of the description intact. This is because users might provide a draft description in the beginning without knowing what are in-scope images look like in the wild. As a result, they might introduce too many details that are not actually not necessary for the concept.
</step1>

<step2>
Based on your response in step1, reason how would you decompose the visual concept into several conditions. Each necessary condition should be concise, self explanatory, and easily understood by human annotators. Each necessary condition should only focus on one concept, and you should not generate a still complex necessary condition. For each condition, you should provide a description of the condition, and a concept name that summarizes the aspect this condition focuses on. Explicitly write down a description and a name for your decomposed conditions here. Remember not all concepts can be further decomposed.
</step2>

<step3>
Examine the following aspect for your decomposition.
(1) Each condition must not significantly overlap with other conditions in their focused concepts.
Examples
- the condition that "a family is gathering together" and "a family is having a meal together" are too similar to each other despite they have different focus.
- the condition that "a vase is broken into pieces" and "someone used some tools to break a vase into pieces" are too similar to each other despite slightly different wording and focus.."
(2) Each condition carries meaningful information, meaning it should not hold true for every image.
For instance, the condition ``the image describes an object'' is too broad and would be true for all images.
</step3>
<step4>
Write down your decomposed conditions.
In cases when you find it hard to decompose the concept, you can just write down a improved concept description. of the original concept here.

You should follow these guidance in your description.
(1) Avoid using verbs, adjectives, or adverbs that carry nuances unless they are an important part of the concept.
Examples
- if the original concept only uses the phrase "show a group of children", then avoid adding using the phrase "depict a group of children" as "depicts" introduces the slight emphasis on visual aspects and ignore the possibility of textual information in the image.
- if the original concept only mentions the phrase "electronic devices", then avoiding using the phrase "such as a phone or a laptop" as the new listed examples suggest a focus on these specific examples.
- if the original concept only uses the phrase "show a beautiful park", then avoid using the phrase "clearly show a beautiful park" because "clearly" implies a degree of visibility to the original concept.

(2) Avoid further defining complex, abstract, or subjective concepts in the original concept definitions.
We only want to break down a composite concept into more unit concepts, and we will define them more clearly in the next round.
Examples:
- if the original concept is "show a beautiful painting", you should just decompose it into "show a painting" and "the painting is beautiful"; you do not need to explicate what make a painting beautiful.
- if the original concept is "show people are gathered happily", you should just decompose it into "show people" and "people are gathered happily"; you do not need to explain visual elements of being happy.

(3) You must not include information beyond the provided information, as new information would effectively change the intended meaning of the concept.
Examples
- if the original concept "woman in a bra" does not mention the context "at the beach", then do not add this context in your necessary conditions.
</step4>

Provide your answer in the following XML structure:
<new-description>Your refined description</new-description>
<reasoning>Add your reasoning here at step2</reasoning>
<examination>Add your examination here at step3</examination>
<conditions>
  <condition>
    <description>Add a condition here</description>
    <name>a short name that summarizes the description</name>
  </condition>
  <condition>
    <description></description>
    <name></name>
  </condition>
  <!-- Add more necessary conditions here if needed. -->
</conditions>

[omit few-shot examples here]

  <visualConceptName>{definition.concept}</visualConceptName>
  <visualConceptDescription>{definition.description}</visualConceptDescription>
\end{promptbox}

%% file: prompt_brainstorm_golden_category_prompt.tex
\begin{promptbox}
<role>You are an expert linguist recognized internationally.</role>
<input>You will be provided an overall concept definition and a focus concept.</input>
<task>
The concept owner wants to catch all images that are in-scope for the focus concept.
To make sure that his decisions about image classifications are consistent, he wants to explicitly clarify the decision-making boundaries for this focus concept.
However, as he starts with a few images, it is possible that he either starts with a narrower concept or a broader concept than he actually wants.
Your task is to infer what are the golden subconcepts that the concept owner wants to include within the scope of this focus concept.
</task>

<step1>
Reason what is the primary concept that the concept owner wants to explicitly define.
There are a few requirements.
1) This description might mention several concepts but you should only focus on the primary concept.
2) If the context indicates that the focus concept is part of the necessary signals of a larger concept, then the primary concepts of these necessary signals should focus on different subconcepts of this larger concept.
In other words, your primary concept should have a different focus than those of the other necessary signals.
3) The primary concept should be more categorical (concepts where you could think about specific instances) rather than descriptive (where you could only describe different aspects of the concept).
Examples of categorical concepts are "fruit", 'electronic devices', 'physical affection', 'outdoor activities',
whereas examples of descriptive concepts are "sleeping person", 'romantic relationship', 'sexual suggestive content'.
</step1>

<step2>
List out categories of subconcepts that have been explored before.
Here the categories mean a way to categorize specific subconcepts, for instance, 'earphones' as a subconcept can belong to the category 'electronic devices' and 'accessories' at the same time.
This includes the following two cases:
1) categories that are already included in the concept definition as positive or negative signals.
2) categories that have beed explored in the previous rounds of brainstorming.
</step2>

<step3>
Based on your answer in step2 and step3, reason and propose a category of subconcepts that you think is the most coherent and widely recognized.
While you can include previous explored subconcepts, your category should not significantly overlap with previously explored categories that you listed in step3
This is because a subconcept can belong to multiple categories at the same time.
In particular, we have the following requirements:
<requirements>
1. You should ensure that this category itself is a well-defined and well-known concept so that average people can easily tell whether an image satisfies this category or not.
2. Your category should not be too narrow that it only covers one or two instances.
3. In cases where there are many potential categories of subconcepts, you should prioritize the one that most people would agree to be in-scope for the concept.
4. You do not aim for proposing a category that includes the most subconcepts; Instead, you should prioritize proposing a category that is coherent, and well-defined.
</requirements>

<examples>
- For the primary concept "fruits", "fruits with red internal flesh" is not a well-known concept, whereas "citrus fruits" is.
- For the primary concept "flowers", "Flowers with five petals" is not a well-known concept and also difficult to recognize in an image; "Rose varieties" is instantly recognizable.
- For the primary concept "birds", "birds of prey" is a well-known concept, visually distinct, and specific without being too narrow.
- For the primary concept "buildings," while "buildings with green roofs" may not be a widely recognized concept,  "religious buildings" is a well-known concept that is often visually distinctive.
- For the primary concept "health care products," while "Vitamin C supplements" might seem like a well-defined category, it is too narrow.
  "Dietary supplements" is a more suitable category encompassing various products like vitamins, minerals, herbs, and fish oil, providing a broader yet well-defined and easily recognizable concept.
</examples>
</step3>

<step4>
For the category in step4, you should write a one-sentence description and a shortname.
Your description should be concise, self explanatory, and easy for an average person to determine if an image satisfies this subconcept or not.

<description-requirements>
You should be careful about your language, in particular, there are several requirements.
1) The recommended format for the description would be "Images show [a general term for the subconcept], such as [at most three specific examples from step3]".
These examples shoulld be representative of the subconcept and should be as specific as possible so that human image annotators can easily know whether an image includes this example or not.
These examples should form a coherent category, and should be specific.
Your short name for the description should be exactly the term that covers this set of examples.

2) Avoid concept descriptions with too many specific and unnecessary details.
    e.g., for the concept 'beverages', your subconcept description should just be 'Images showing various types of tea drinks such as green tea, black tea, and herbal tea'
    rather than 'Images that show people drinking various types of tea drinks with different colors and flavors such as green tea, black tea, and herbal tea'
  Similarly, avoid adding too many unnecessary details to the examples you provide.
  For instance, "eagles" is a good example of the category "birds of prey", but "bald eagles" or "eagles in the sky" are not.

3) Be careful about your word choices of verbs, nouns, or adjectives, which might carry unexpected nuances.
    e.g., be careful about using 'depict' or 'mention', or 'show' as the previous two verbs introduce the slight emphasis on visual or textual aspects.
    e.g., be careful about using adjectives like 'clearly' or 'explicitly' as they might suggest a degree of visibility to the original concept.
</description-requirements>
</step4>

Write your output strictly in this valid xml format:
  <repeat-focus-concept>Repeat the focus concept here.</repeat-focus-concept>
  <primary-concept>List out the primary concept this categorical concept focuses on here.</primary-concept>
  <explored-subconcepts>
  List out the questions that have been asked for this concept before if any in a bullet point list at step2.
  </explored-subconcepts>
  <category-reasoning>
  List out your reasonings at step3 here.
  </category-reasoning>
  <subconcept>
    <description></description>
    <name></name>
  </subconcept>

[omit few-shot examples here]

<conceptDefinition>
  {str(definition)}
</conceptDefinition>
<previous-signals>
  {previous_signal_str}
</previous-signals>s
<context>{context}</context>
\end{promptbox}

%% file: prompt_brainstorm_borderline_category_prompt.tex
\begin{promptbox}
<role>You are an expert linguist recognized internationally.</role>
<input>You will be provided a concept definition and an optional context for this concept.</input>
<task>
The concept owner wants to catch all images that are in-scope for this concept.
To make sure that his decisions about image classifications are consistent, he wants to explicitly clarify the decision-making boundaries for this concept.
However, as he starts with a few images, it is possible that he either starts with a narrower concept.
Your task is to suggest a new borderline subconcept that the concept owner might want to include.
</task>

<step1>
Reason what is the primary concept that the concept owner wants to explicitly define.
There are a few requirements.
1) This description might mention several concepts but you should only focus on the primary concept.
2) If the context indicates that the focus concept is part of the necessary signals of a larger concept, then the primary concepts of these necessary signals should focus on different subconcepts of this larger concept.
In other words, your primary concept should have a different focus than those of the other necessary signals.
3) The primary concept should be more categorical (concepts where you could think about specific instances) rather than descriptive (where you could only describe different aspects of the concept).
Examples of categorical concepts are "fruit", 'electronic devices', 'physical affection', 'outdoor activities',
whereas examples of descriptive concepts are "sleeping person", 'romantic relationship', 'sexual suggestive content'.
</step1>

<step2>
If the user starts with a narrower concept, reason what will be the broader concept that the user might want to define.
You should examine the given context and summarize the broader concept the user might intend to say in replacement of the primary concept.
This broader concept should fit nicely with other parts of the context.

<example>For the concept 'health supplements' within the context of "images that show health supplements to promote wellness", the broader concept might be 'wellness products'</example>
<example>For the concept 'fake websites' within the context of "images that show fake websites as part of online fraud", the broader concept might be 'online fraudulent schemes'</example>
<example>For the concept 'electronic devices' within the context of "images that show electronic devices in a library", the broader concept might be 'things we can find in a library'</example>
</step2>

<step3>
List out categories of edgecase categories that have been explored before from the previous signals input.
</step3>

<step4>
Based on your answer in step1 and step2, what other subconcepts might be in-scope for this broader concept in step2 but obviously not part of the primary concept in step1.
In other words, you should not focus on detailing specific edgecase categories of this primary concept. Instead you should think about what other subconcepts (replacing the primary concept) frequently appear in the same context.

There are a few requirements.
<requirements>
1) You should NOT focus on detailing specific edgecase categories of this primary concept.
2) Your category should NOT significantly overlap with the subconcepts that have been explored at step3.
3) Your category should not refer to examples that significantly overlap with the examples that have been explored before in step2.
4) We will later define the other necessary signals for this concept, so your category should NOT try to define other necessary signals.
</requirements>

<example>For the concept 'health supplements' within the context of "images that show health supplements to promote wellness", 'fresh fruits', 'yoga mats', or 'spa treatments' might also be interesting because they can also visually symbolize natural and holistic approaches to health and well-being.</example>
<example>For the concept 'fake websites' within the context of "images that show fake websites as part of online fraud", "counterfeit product pages", "fake social media accounts", or "fraudulent payment forms" might also be interesting because they can also be depicted as part of online fraud schemes.</example>
<example>For the concept 'electronic devices' within the context of "images that show electronic devices in a library", 'board games', 'books', or 'musical instruments' might also be interesting because they might also appear in a library despite not electronic.</example>
</step4>

<step4>
Based on your reasoning in step4, write a one-sentence description and a shortname for your borderline subconcepts.

<description-requirements>
You should be careful about your language, in particular, there are several requirements.
1) The recommended format for the description would be "Images show [a general term for the subconcept], such as [at most three specific examples from step3]".
These examples shoulld be representative of the subconcept and should be as specific as possible so that human image annotators can easily know whether an image includes this example or not.
These examples should form a coherent category, and should be specific.
Your short name for the description should be exactly the term that covers this set of examples.

2) Avoid concept descriptions with too many specific and unnecessary details.
    e.g., for the concept 'beverages', your subconcept description should just be 'Images showing various types of tea drinks such as green tea, black tea, and herbal tea'
    rather than 'Images that show people drinking various types of tea drinks with different colors and flavors such as green tea, black tea, and herbal tea'
  Similarly, avoid adding too many unnecessary details to the examples you provide.
  For instance, "eagles" is a good example of the category "birds of prey", but "bald eagles" or "eagles in the sky" are not.

3) Be careful about your word choices of verbs, nouns, or adjectives, which might carry unexpected nuances.
    e.g., be careful about using 'depict' or 'mention', or 'show' as the previous two verbs introduce the slight emphasis on visual or textual aspects.
    e.g., be careful about using adjectives like 'clearly' or 'explicitly' as they might suggest a degree of visibility to the original concept.
</description-requirements>
</step4>

Write your output strictly in this valid xml format:
  <primary-concept>List out the primary concept this categorical concept focuses on here.</primary-concept>
  <broader-concept>List out the broader concept that the user might want to define here.</broader-concept>
  <previous-signals>List out the previous signals explored before here at step3 in a bullet point list.</previous-signals>
  <reasoning>
  List out your reasonings at step4 here.
  </reasoning>
  <subconcept>
    <description></description>
    <name></name>
  </subconcept>

<conceptDefinition>
  {str(definition)}
</conceptDefinition>
<previous-signals>
{previous_signal_str}
</previous-signals>
<context>{context}</context>
\end{promptbox}

%% file: prompt_generate_descriptions_prompt.tex
\begin{promptbox}
<role>You are an expert linguist who is good at brainstorming creatively.</role>
<input>
You are given a structured concept definition and a list of previously generated descriptions.
</input>
<task>
Your task is to generate {num_descriptions} more description that covers a possibly {image_type} category of images.
This category of images should be different from the categories covered by previous descriptions.
Since users still explore and improve their concept definition, you should also consider similar categories
that might be fully covered by the current definition but are relevant.
This generated description will later be used to find images that satisfy the concept through a search engine.
</task>

<define-a-concept>
Before answering the question, it is a must for you to understand how we define a concept in a structured and iterative way.
- If the concept is defined by a list of necessary conditions, then an image is in-scope if it satisfies all necessary conditions.
- If the concept is defined by a list of positive and negative conditions, then an image is in-scope if it satisfies at least one positive condition and does not satisfy any negative condition.
- A concept can be first defined by a list of necessary conditions, and then each necessary condition can be further defined by a list of positive and negative conditions.
</define-a-concept>
<image-type>
- in-scope: images that are likely to be in-scope for the concept.
- ambiguous: images that are borderline in-scope for the concept. 
There are some important ambiguities that the concept definition does not articulate clearly.
- out-of-scope: images that are likely to be out-of-scope for the concept.
</image-type>

<step1>
Examine the concept definition and previous descriptions, propose {num_descriptions} new categoriesof images are in-scope but are different from the previous descriptions.
These categories should cover significantly different categories of images from the previous descriptions.
</step1>

<step2>
Based on your reasoning in step1, write down {num_descriptions} new descriptions for the {image_type} category of images.
The description should be concise, specific, and clear. You should aim for less than 20 words for this description.
</step2>

Write your answer in a valid XML format, adhering to the following structure:
<reasoning>
  Write your reasoning for new categories of images in step 1 here.
</reasoning>
<descriptions>
  Write your description for the in-scope category in step 2 here.
  <!-- there should be {num_descriptions} descriptions in total -->
  <description></description>
  <description></description>
  <description></description>
</descriptions>

Here is the concept definition you should work on:
<definition>{definition.readable_string()}</definition>
<previous-descriptions>
{previous_descriptions_str}
</previous-descriptions>
\end{promptbox}

%% file: prompt_whether_borderline_prompt.tex
\begin{promptbox}
<role>You are an expert linguist who is good at brainstorming creatively.</role>
<input>
You are given the definition of a visual concept,
which serves as the guideline for human raters to determine whether an image is within the scope of this concept.
You will also be given an image.
</input>
<task>
As the concept owner is still actively working on the definition of the target concept,
there are still some ambiguities in this concept definition.
You will help examine whether an image might highlight important ambiguities in the concept definition and thus should be reviewed by the concept owner,
so that the concept owner could further improve the definition.
</task>

<step1>
Examine the image against the definition and determines whether the image should be classified as in-scope or out-of-scope.
</step1>
<step2>
Now assume that the concept owner actually gives a different classification result for this image.
Examine the image against the definition and reason what might be the important ambiguities that the current definition fails to capture.
As the current definition is mostly correct, you should not completely ignore the current definition,
but instead you should focus on identifying subtle but important ambiguities.
</step2>
<step3>
Examine your reasoning in step2, and determine how likely these ambiguities actually make sense.
This means that the concept owner is likely to be unclear about his definition at this point, or this point is likely to cause confusion to human raters.
Some images are actually clear-cut in-scope or out-of-scope examples--in these cases, your ambiguities might not make much sense.
</step3>
<step4>
If you believe that the ambiguities are important in step2, pick the most important ambiguity from step2 and summarize it in one sentence less than 30 words.
Your summary should directly point out the elements that might cause the ambiguity in the image and the specific requirements in the definition.
for instance,
- "The image shows two people use sign language to communicate with each other, but it is unclear whether sign language is considered as "chatting"", 
- "The image shows a set of cartoon dogs, but it is unclear whether cartoon is considered as "dog" or not."
But if you believe that the ambiguities are not important, then your summary should be an empty string.
</step4>

Provide your answer in the following XML structure:
<classification>The classification result of the image and your reasoning.</classification>
<counter-reasoning>Your reasoning about why the image might have been misclassified at step2</counter-reasoning>
<examination>Your reasoning about whether the ambiguities are important at step3</examination>
<summary>Your summary of the most important ambiguity if your answer is "yes" at step3; otherwise, leave it empty.</summary>

Here is the definition you should work on:
<visualConceptDescription>{definition.readable_string()}</visualConceptDescription>
\end{promptbox}

%% file: prompt_reflect_classified_image_prompt.tex
\begin{promptbox}
<role>You are an expert linguist</role>.
<input>
  You are given the definition of a visual concept and an image caption.
  Guided by this concept definition, human raters are asked to determine whether an image is within the scope of this concept, and write down their rationales and decisions.
  We also asked the concept owner to directly rate whether this image is in-scope or out-of-scope of this concept [ground-truth]
  The concept owner might also provide feedback regarding what the human raters should have noticed.
</input>

<task>
  The concept owner is still actively working on the definition of the target concept.
  The final goal is to enable human raters to interpret this definition to rate images exactly as the concept owner would do.
  Your task is to help articulate what this concept owner wants to clarify for this visual concept.
</task>

<define-a-concept>
Before answering the question, it is a must for you to understand how we define a concept in a structured and iterative way.
- If the concept is defined by a list of necessary conditions, then an image is in-scope if it satisfies all necessary conditions.
- If the concept is defined by a list of positive and negative conditions, then an image is in-scope if it satisfies at least one positive condition and does not satisfy any negative condition.
- A concept can be first defined by a list of necessary conditions, and then each necessary condition can be further defined by a list of positive and negative conditions.
</define-a-concept>

<step1>
Within the context of this image, reason over what clarifications the concept owner might want to incorporate into the definition.
You should try to refer to specific elements in the image to better ground your reasoning.
Your reasoning should be based on the following questions:
1) If the concept owner provides a clear feedback, what do you think the concept owner wants to clarify? 
  Do not generalize too much beyond what the concept owner says.
2) If the concept owner provides a different rating than the human raters, what is the possible reason for this disagreement?
  Do not generalize too much beyond this disagreement between ratings.
3) When the concept owner provides no clear feedback and the human raters and the concept owner are in agreement, 
  what does this agreement between the concept owner and the human raters confirm?
  Especially in this scenario, since there is no less clear information, 
  you should be more conservative and specific, and try to avoid generalizing too much.
</step1>
<step2>
Summarize your reasoning in the step 1 with a few sentences.
Your summary will be used in downstream steps so make sure it covers all the necessary information.
Please make sure that you ground your summary with specific elements or examples in the image.
this will help others better understand your reasoning.
</step2>

Provide your answer in a valid XML format, adhering to the following structure:
  <reasoning>Describe your reasoning in the step 1</reasoning>
  <clarification>Provide your answer to the question in the step 2</clarification>

<conceptDefinition>{definition.readable_string()}</conceptDefinition>
<raterResponses>
  <decision>{ImageClassifier.rating_to_label(reflection_info['decision'])}</decision>
  <summary>{reflection_info['summary']}</summary>
</raterResponses>
<conceptOwner>
  <ground-truth>{ImageClassifier.rating_to_label(reflection_info['groundtruth'])}</ground-truth>
  <user-feedback>{reflection_info['feedback']}</user-feedback>
</conceptOwner>
<image_caption>{image.image_caption}</image_caption>
\end{promptbox}

%% file: prompt_improve_definition_prompt.tex
\begin{promptbox}
<role>You are an expert linguist</role>.
<input>
  You are given the definition of a visual concept, which serves as the guideline for human raters to determine whether an image is within the scope of this concept.
  However, as the concept owner is still actively working on the definition of the target concept, human raters incorrectly rated an image regarding a particular focus concept within this definition.
  Therefore, the concept owner wants to clarify their points and improve the definition of this focus concept.
  They provide their clarifications for each of {images_num} images respectively.
  The final goal is to have a more accurate and comprehensive concept definition so that human raters can rate images exactly as the concept owner would do.
</input>

<task>
  Your task is to determine how to improve the definition of the focus concept in the most accurate and concise way.
</task>

<define-a-concept>
Before answering the question, it is a must for you to understand how we define a concept in a structured and iterative way.
- If the concept is defined by a list of necessary conditions, then an image is in-scope if it satisfies all necessary conditions.
- If the concept is defined by a list of positive and negative conditions, then an image is in-scope if it satisfies at least one positive condition and does not satisfy any negative condition.
- A concept can be first defined by a list of necessary conditions, and then each necessary condition can be further defined by a list of positive and negative conditions.
</define-a-concept>

<step1>
Examine the clarifications for all these images and summarize the key points that the concept owner might want to incorporate into the definition.
Some clarifications might be similar, so you should aggregate them;
On the other hand, some clarifications might be very different, so you should consider listing them separately.
</step1>
<step2>
Based on your answers in previous steps, reason how to incorporate the key points into the definition.
You could either choose to add a new positive or negative signal to the definition, or modify an existing positive or negative signal.
For an existing positive or negative signal, you should not add a child positive or negative signal to it.
For an existing necessary condition, you should not add a sibling positive/negative/necessary signal to it.
</step2>
<step3>
Based on your answers in previous steps, write down the changes you want to make to the definition.

You should be careful about the language of your changes, in particular, there are several requirements.
<description-requirements>
  1) Always make sure that your final description is CONCISE, COHERENT, and ACCURATE; an average person could easiy determine whether an image satisfies the signal based on the description.
  2) DO NOT write a complex sentence structure in a description of a signal.
  3) You should only make important changes to the description.

  If the original description misses a point, you are encouraged to use one of the following ways to incorporate the nuances the concept owner wants to convey:
  a) add new adjectives, b) use different verbs, or c) add a few constraint words.
  If the original description uses an ambiguous or misleading word or example, you are encouraged to a) refine the word or example, or b) simply remove them from the description.

  4) Be careful about your word choices of verbs, nouns, or adjectives, which might carry unexpected nuances.
      e.g., be careful about using 'depict' or 'mention', or 'show' as the previous two verbs introduce the slight emphasis on visual or textual aspects.
      e.g., be careful about using adjectives like 'clearly' or 'explicitly' as they might suggest a degree of visibility to the original concept.
  5) If your description consists of two independent conditions, you might consider use the format like "Images that 1) ... and 2) ..." to make it more clear.
</description-requirements>

</step3>

Provide your answer in a valid XML format, adhering to the following structure:
  <keypoints>Describe your reasoning of the key points of these clarifications in the step1</keypoints>
  <reasoning>Describe your reasoning of how to incorporate the key points into the definition in the step2</reasoning>
  <improve-description>
    The improved description of the visual concept in the step3.
    You should only write down changes you proposed in the following format.
    1) If you want to edit an existing signal, the format is as follows:
    <concept>
      <name>The name of the signal you want to edit</old-name>
      <old-description>The original description of the signal</old-description>
      <new-description>The new description of the signal</new-description>
    </concept>
    2) If you want to add a new signal, the format is as follows:
    <concept>
      <parent-signal>The name of the parent signal</parent-signal>
      <type>The type of the new signal, either 'positive' or 'negative'</type>
      <new-name>The new name of the signal</name>
      <new-description>The new description of the signal</description>
    </concept>
    It might be possible that you need to make multiple changes, so you should write down all of them. 
  </improve-description>

<conceptDefinition>{definition.print_definition()}</conceptDefinition>
<clarifications>{reflections_str}</clarifications>
\end{promptbox}